\documentclass{article}

\usepackage{lipsum}
\usepackage{graphicx}
\usepackage{algorithm}
\usepackage[noend]{algpseudocode}
\usepackage{enumerate}
\usepackage{url}
\usepackage{pgfplots}
\usepackage{soul}
\usepackage{subcaption}
\usepackage{xr}
\usepackage{multirow}

\newcommand{\erdos}{Erd\H{o}s-R\'enyi}
\newcommand{\barabasi}{Barab\'asi-Albert}
\newcommand{\strogratz}{Watts--Strogatz}

\title{GoT-WAVE: Temporal network alignment using graphlet-orbit transitions}
\author{David Apar\'icio, Pedro Ribeiro and Fernando Silva \vspace{0.1cm}\\
	CRACS \& INESC-TEC, \\Faculdade de Ci\^encias, Universidade do Porto, \\ R. Campo Alegre, 1021, 4169-007 Porto, Portugal. \\
	\vspace{-0.1cm}
	\and
	Tijana Milenkovi\'c  \vspace{0.1cm}\\
	Department of Computer Science and Engineering, \\Interdisciplinary Center for Network Science and Applications,\\ and ECK
	Institute for Global Health, \\University of Notre Dame, \\Notre Dame, IN 46556, USA.  \\
}

\date{\today}
\begin{document}
	
	\maketitle

	\begin{abstract}
		 Global pairwise network
	alignment (GPNA) aims to find a one-to-one node mapping
	between two networks that identifies conserved network
	regions. GPNA algorithms optimize node conservation (NC) and
	edge conservation (EC). NC quantifies topological similarity
	between nodes. Graphlet-based degree vectors (GDVs) are a
	state-of-the-art topological NC measure. Dynamic GDVs (DGDVs)
	were used as a dynamic NC measure within the first-ever
	algorithms for GPNA of \emph{temporal} networks: DynaMAGNA++
	and DynaWAVE. The latter is superior for larger networks.  We
	recently developed a different graphlet-based measure of
	temporal node similarity, graphlet-orbit transitions
	(GoTs). Here, we use GoTs instead of DGDVs as a new dynamic NC
	measure within DynaWAVE, resulting in a new approach,
	GoT-WAVE. 
	
	On synthetic networks, GoT-WAVE
	improves DynaWAVE's accuracy by 25\% and speed by 64\%.  On
	real networks, 
	when optimizing only dynamic NC, each method is superior
	$\approx$50\% of the
	time. While DynaWAVE benefits more from also optimizing
	dynamic EC, only GoT-WAVE can support \emph{directed} edges.
	Hence, GoT-WAVE is a promising new temporal GPNA algorithm,
	which efficiently optimizes dynamic NC. Future work on better
	incorporating dynamic EC may yield further improvements.\\
	
	\end{abstract}

\section{Introduction}
	
	\vspace{0.2cm}
	
Network alignment (NA) aims to find  similar (conserved)
regions between compared networks. These regions are not expected to be perfect fits, and thus, NA deviates from the
traditional subgraph isomorphism problem
\cite{ullmann1976algorithm}. 
Then, NA can be used
for knowledge transfer from a well-known
system to a poorly-studied
system between their conserved network regions ~\cite{elmsallati2016global}. For example, in computational biology, NA can be used to identify topologically similar (and possibly also sequence-similar) regions of molecular networks of different species and to predict functions of currently unannotated proteins based on functions of their aligned partners in another network \cite{faisal2015global}. 

NA produces either: (a) a many-to-many mapping of highly
conserved but small network regions or (b) a one-to-one
mapping that covers every node of the smaller network and equally many nodes from the other network and is thus large, but
 is often suboptimally conserved
\cite{elmsallati2016global,faisal2015post}.
Both NA types, called local and global,
respectively, have (dis)advantages
\cite{meng2016local,guzzi2017survey}. We
focus on global NA. 

Global NA can
be pairwise~\cite{klau2009new}, resulting in aligned pairs
of nodes between two networks, or
multiple~\cite{flannick2008automatic,vijayan2017multiple}, resulting in
aligned node clusters between three or more networks. Multiple NA is
more computationally complex than pairwise NA and, furthermore, recent
work suggests that multiple NA is also less accurate than pairwise
NA~\cite{vijayan2017pairwise}. So, here, we focus on pairwise
NA, and in particular on global pairwise NA (GPNA).

GPNA consists of two algorithmic components: 1) an objective function,
typically node conservation (a measure of node similarity) combined
with edge conservation, and 2) an optimization strategy (also called
alignment strategy) that aims to maximize the objective
function. 

Regarding the first component and specifically the node conservation part,
graphlet degree vectors
(GDVs)~\cite{prvzulj2007biological,milenkovic2008uncovering}
have been widely used as topological properties (features) to measure
node conservation in GPNA due to the rich topological information that
graphlets
capture~\cite{milenkovic2010optimal,memivsevic2012c,vijayan2015magna++}. GDV-based
node conservation was shown to be superior in the task of GPNA under
the same optimization strategy to other node conservation/similarity
measures: IsoRank's PageRank and GHOST's spectral
signature measures from the biological domain
\cite{faisal2015global,crawford2015fair}, or node2vec and struc2vec network
embedding measures from the social domain
\cite{gu2018graphlets}.
Regarding the first component and specifically the edge
conservation part, several established measures of edge conservation
exist: S$^3$, which rewards an alignment when edges are aligned to each other and penalizes
it when  an edge is aligned to a non-edge~\cite{saraph2014magna}, and weighted edge
conservation (WEC), which is high if many edges are aligned to each other
\emph{and} the nodes of the aligned edges are  similar with respect to
 node conservation \cite{sun2015simultaneous}.

Regarding the second component, existing  GPNA
algorithms have one of two types of optimization
strategy. One type is seed-and-extend, where
first two highly similar nodes (with respect to node conservation) are aligned, i.e., seeded.  Then, the seed's network
neighbors that are similar are aligned, the seed's neighbor's
neighbors that are similar are aligned, and so on.  The extension
around the seed and exploration of the seed's neighbors aims to
improve both node and edge conservation of the resulting alignment.
The extension continues until all nodes in the smaller  network are aligned, i.e., until a one-to-one (injective) node mapping
is produced.  WAVE is a representative state-of-the-art
seed-and-extend optimization strategy (that we focus on for reasons
discussed below), which by default optimizes GDV-based node
conservation and WEC
\cite{sun2015simultaneous}. The other type of optimization strategy
is a search algorithm.  Here, instead of aligning node by node as with the seed-and-extend approach, entire alignments are
explored and the one with the best  objective
function score is returned. MAGNA++
\cite{vijayan2015magna++} is a representative  state-of-the-art 
search algorithm (that we focus on for reasons discussed below), which
uses a genetic algorithm to, by default, optimize GDV-based node
conservation and S$^3$. Importantly, a typical
optimization strategy, including WAVE and MAGNA++, can 
optimize any objective function, i.e., it is not limited to e.g., GDV-based node conservation and S$^3$ or WEC.
Note that in a recent comprehensive evaluation of different methods \cite{meng2016local},WAVE and MAGNA++ rose to the top, although  newer GPNA methods have appeared since, such as SANA \cite{mamano2017sana}. 
	
Traditional GPNA methods align \emph{static}
networks~\cite{vijayan2017pairwise}.  However, because most of
real-world systems evolve over time and thus exhibit a dynamic nature,
they are intrinsically not static.  As such, they can only be truly
understood by accounting for their
evolution~\cite{holme2012temporal}. The first-ever methods for GPNA
of \emph{temporal} networks (see below) were proposed only 
recently. This could be due to limitations of current biotechnologies
for data collection, which have resulted in a lack of temporal network
data on molecular systems, such as protein-interaction networks
(PINs), that are the systems to which static GPNA methods have been
extensively
applied~\cite{kelley2003conserved,milenkovic2010optimal,vijayan2015magna++}.
However, as initial temporal PIN data begin to emerge
~\cite{faisal2014dynamic,yoo2015improving}, and as other temporal
network data become available, e.g., brain, ecological, or social networks~\cite{gao2011temporal,sociopatterns,olesen2008temporal,rossidr},
temporal GPNA will gain increasing importance.

The only temporal GPNA methods currently available are
DynaMAGNA++ \cite{vijayan2017alignment} and
DynaWAVE~\cite{vijayan2017aligning},  temporal extensions of MAGNA++ and WAVE.  DynaWAVE was shown to be
more accurate and faster than DynaMAGNA++ on medium- and large-size
networks; DynaMAGNA++ was more accurate (yet slower) on
small-size ($\approx$100-node) networks. Since most of
real-world networks are not small, we focus on
DynaWAVE. This method uses the same seed-and-extend optimization
strategy as static WAVE, but it uses it to optimize \emph{dynamic} node and edge conservation. As its dynamic node
conservation, DynaWAVE uses a temporal extension of GDVs, 
dynamic GDVs (DGDVs), which were originally proposed for tasks
of node and network classification
by~\cite{hulovatyy2015exploring}. DGDV of a node uses
dynamic graphlets to describe the node's neighborhood in a
temporal network. Comparing nodes' DGDVs yields a measure
of similarity between the nodes' evolving neighborhoods, i.e.,
dynamic node conservation. As its dynamic edge conservation, DynaWAVE
uses dynamic WEC (DWEC), a temporal analog of
WAVE's WEC that generalizes an aligned edge to
an aligned event (temporal edge) 
~\cite{vijayan2017aligning}. Just as WAVE, DynaWAVE can use its optimization strategy in combination with any objective function.

We recently developed graphlet-orbit transitions
(GoTs)~\cite{aparicio2017temporal}, a different temporal
graphlet measure of node similarity. GoTs describe 
how a node's neighborhood is evolving by measuring how its participation in different graphlet positions (orbits) changes with time. For example, GoTs can capture when a node in the center of a $k$-node star at time $t$ becomes a part of a $k$-node clique at time $t+1$. So far, we used GoTs for network
classification. Here, we aim to use GoTs for temporal GPNA as a new dynamic node conservation measure within DynaWAVE. We refer
to our GoT-modified version of DynaWAVE as GoT-WAVE. 

We evaluate whether GoT-WAVE improves upon DynaWAVE by
mimicking the evaluation from the DynaWAVE
study. Namely, we evaluate on synthetic data containing 50 temporal networks produced by dynamic versions of five
well known graph models. Here, we align all pairs of networks to each
other. A good temporal GPNA method should identify as similar those
networks that originate from the same model and as dissimilar those
networks that originate from different models.  Also, we compare the methods on eight real-world networks from biological and social
domains. Here, we align each network to its noisy
version, in which a percentage of the original network's
edges is rewired. Since the aligned networks have the same
nodes, we know which nodes should be mapped to which nodes.  The more nodes are correctly mapped, the better the
method. In all evaluation
tests, we compare the two methods when they optimize: 1) only their
respective dynamic node conservation measures (GoTs versus DGDVs), to
\emph{fairly} evaluate the two measures against each other, and 2)
both node and edge conservation, to give each method the
\emph{best-case} advantage (it was already shown that DynaWAVE performs better when it optimizes both  rather than only one of node and edge conservation).

We find that on synthetic networks, under both the
fair and best-case scenario, GoT-WAVE is
 more accurate than DynaWAVE by  25\% and  faster by 64\%.  On real networks,
under the fair scenario, GoT-WAVE is more
accurate that DynaWAVE for four of the eight networks, performing
better on the denser networks and worse on the sparser ones. We observe the opposite in terms of their running times, i.e., GoT-WAVE is slower than DynaWAVE for denser networks and faster for sparser networks.  Thus, the
two methods are complementary. Under the best-case scenario, DynaWAVE's performance is  more
enhanced than GoT-WAVE's when dynamic edge conservation is considered
as well, as DynaWAVE is now better for all
eight networks. However, because GoTs is the only current
temporal graphlet-based measure of node similarity that supports edge
direction, GoT-WAVE is the only temporal GPNA method that can deal
with directed networks (DGDVs and thus DynaWAVE always assume that edges are undirected).

Thus, GoT-WAVE is a promising new
temporal GPNA method that efficiently optimizes dynamic node conservation.  Finding new measures of dynamic edge conservation better suited for GoT-WAVE could further enhance its performance, which is the subject of future work. Also, GoTs, when used as node conservation within any newer or future GPNA optimization strategies, such as SANA, could yield further improvements. 

	\section{Methods}\label{sec:netwligngot}
	
	
	\subsection{Concepts and terminology}\label{sec:terminology}
	
	A network or \textit{graph} $G$ is comprised of a set of \textit{vertices} or \textit{nodes}, $V(G)$, and a set of \textit{edges}, $E(G)$. Nodes represent entities and edges correspond to relationships between the entities. Edges are represented as pairs of vertices of the form $(a, b)$, where $a, b \in V(G)$. In \textit{directed} graphs, edges $(i, j)$ are \textit{ordered pairs} (translated to ``$i$ \textit{goes to} $j$''), whereas in \textit{undirected} graphs there is no order since nodes are always reciprocally connected. Our proposed methodology (see below) is applicable to both undirected and directed graphs~\cite{aparicio2017extending}. 	A temporal network comprises $s$
	consecutive network snapshots. We denote the set of all snapshots of
	a temporal network $G$ by $\mathcal{S}(G)$, and the $i^{th}$ snapshot  as $\mathcal{S}_i(G)$. A temporal network can gain/lose nodes/edges from
	$\mathcal{S}_i(G)$ to $\mathcal{S}_{i+1}(G)$. We denote the number of nodes in  snapshot $i$ by $N_i$.
	
	\begin{figure}[!t]
		\centering
		\includegraphics[width=0.4\linewidth]{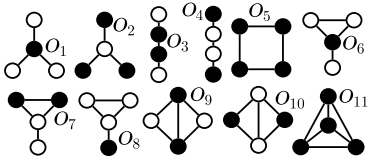}
		\vspace{-0.2cm}
		\caption{All 11 orbits of all six undirected 4-node graphlets. Nodes that are in the same orbit  (i.e., are topologically equivalent) in a given graphlet are colored in black. For example, $o_1$ and $o_2$ are two possible orbits of a 4-node star, orbits $o_3$ and $o_4$ are two possible orbits of a 4-node chain, etc.}
		\label{fig:orbits_u4}
	\end{figure}

\begin{figure}[!h]
		\centering
		\includegraphics[width=0.5\linewidth]{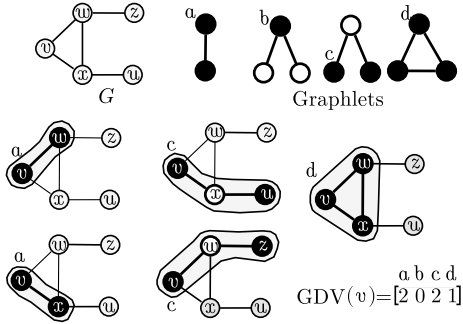}
		\vspace{-0.2cm}
		\caption{Illustration of GDV($v$) that counts how many times $v$ participates in each of the orbits $a$, $b$, $c$ and $d$ of all undirected 2-3-node graphlets. In this example, $v$ touches orbit $a$ twice (i.e., has degree 	of two), the periphery of a 3-node chain (orbit $c$) twice, and a triangle (orbit $d$) once. }
		\label{fig:orbit_occ}
	\end{figure}

	Graphlets are small non-isomorphic subgraphs. Different node positions (or symmetry groups) in a graphlet are called orbits.  For example, it is  different to be at the center of a star or at its periphery. All orbits of all undirected graphlets with four nodes are illustrated in Figure~\ref{fig:orbits_u4}. Graphlets are a general concept  (e.g., not specific to a given size, and edge direction can be incorporated). We denote by $u\mathcal{O}_k$ the set of all orbits of all $k$-node undirected graphlets, and by $d\mathcal{O}_k$  the equivalent for directed graphlets. We use the simpler $\mathcal{O}$ notation when the concept is general. The GDV of node $v$,  GDV($v$), counts how many times node $v$ appears in (i.e., touches) each orbit $o \in \mathcal{O}$ (Figure~\ref{fig:orbit_occ}). The notion of graphlets and GDVs has been extended from the static to  temporal context, into dynamic graphlets and DGDVs. For details, see \cite{hulovatyy2015exploring}.
	
	\subsection{Static and temporal GPNA}
	
	Static GPNA produces an injection \mbox{$f: V(G) \rightarrow V(H)$}, where $V(G)$ is not bigger than $V(H)$, maximizing node or edge conservation between aligned node pairs. Temporal GPNA, extending static GPNA, aims to optimize dynamic node or edge conservation. As dynamic node conservation, we optimize similarity between nodes' temporal graphlet-based features, namely GoTs, which we define in Section~\ref{sec:got}. We compare nodes' GoTs as described in Section~\ref{sec:gotalign}.
	As dynamic edge conservation, when we also optimize this measure, we use DWEC, just as DynaWAVE does. That is, compared to DynaWAVE, the only aspect that we modify is its DGV-based dynamic node conservation measure, replacing it with our GoT-based measure. This ensures a fair comparison between GoTs and DGDVs as two different temporal graphlet-based node features.
	%


	\subsection{Graphlet orbit-transitions (GoTs)}\label{sec:got}
	
	\begin{figure*}[!b]
		\centering
		\includegraphics[width=1\linewidth]{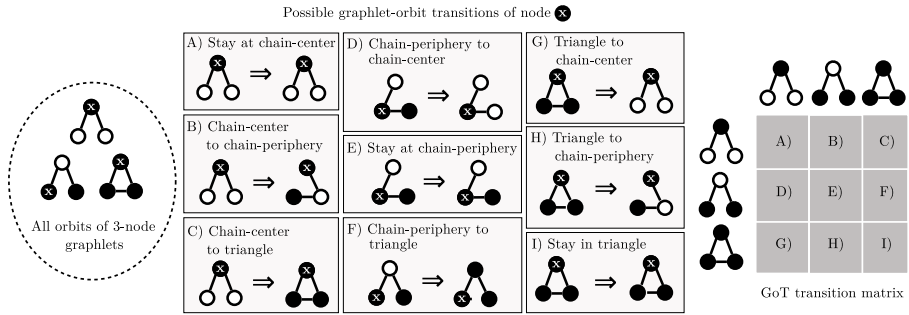}
		\vspace{-0.5cm}
		\caption{All possible graphlet-orbit transitions (GoTs) of 3-node undirected graphlets and the corresponding GoT matrix. Node $x$ is the node being  considered (whose GoT matrix is shown), and black nodes are in the same orbit as $x$. Each cell $(i,j)$ of the GoT matrix represents the number of times node $x$ transitions from orbit $i$ to orbit $j$.}
		\label{fig:fulltrans}
	\end{figure*}
	
	
	Just like DGDVs, GoTs only account for connected graphlets.
	%
	Consider the two possible 3-node undirected connected graphlets (chain and triangle) and their orbits from Figure~\ref{fig:fulltrans}. A chain  has two possible orbits, i.e., a node can be either at the center of the chain or in one
	of its leaves. A triangle has a single orbit, as  all of its nodes are topologically equivalent. GoTs are the matrix
	of changes (transitions) between  every possible pair of orbits across two consecutive snapshots. There are a total of
	$3\times3 = 9$ possible orbit transitions in  Figure~\ref{fig:fulltrans}. A node can
	remain in its previous orbit, be it a (A) chain-center, (D) chain-periphery	or (I) triangle-node. Or, it can transition from the chain-center to the chain-periphery (B)
	or to a triangle-node (C), etc. All possibilities for the 3-node graphlets are illustrated in Figure~\ref{fig:fulltrans}; in practice, we use larger graphlets as well (see below). The matrix from Figure~\ref{fig:fulltrans} illustrates the GoTs of node $x$ that we use as $x$'s feature vector. This matrix offers rich topological information that can be used for various tasks~\cite{aparicio2017temporal}. Here, we use it in the task of temporal GPNA.
	
	Regarding the considered graphlet size, \cite{hulovatyy2015exploring} recommended the use of all DGDVs with up to four nodes and six events (temporal edges). This is what we do,  to give the best-case advantage to DynaWAVE. For a fair comparison, to account for as similar as possible amount of network topology with both DGDVs and GoTs, we also use all undirected GoTs with up to four nodes, unless explicitly stated otherwise.
	
	
	\subsection{GoT-WAVE}\label{sec:gotalign}
	
	For each node, we compute its GoT matrix, flatten the matrix to a vector, and use the vector as the node's features. The feature vectors over all nodes in a network form a $\#Nodes \times \#Transitions$ matrix. For two networks being aligned, this results in two corresponding matrices with the same number of columns, whose rows are then joined together. Due to high dimensionality and sparsity of the  joined matrix, we perform dimensionality reduction on the matrix using principal component analysis, keeping 99\% of its variance. Then, we compute the topological similarity between every two nodes from different networks as the cosine similarity between the nodes' PCA-reduced feature vectors. GoT-WAVE uses the resulting node similarities as the dynamic node conservation part of the objective function, which is then optimized using WAVE. In all of the above steps, we do exactly what DynaWAVE does to produce DGDV-based node similarities and perform DGDV-based temporal GPNA.   
	
	GoT-WAVE, like DynaWAVE, can optimize dynamic node conservation (i.e., GoTs for GoT-WAVE, DGDVs for DynaWAVE), dynamic edge conservation (DWEC), or both. Its objective function is $\alpha S_E + (1 - \alpha)S_N$, where $S_E$ and $S_N$  are dynamic edge and node conservation measures, respectively, and $\alpha \in [0,1]$ controls how important each measure is. We use: 1) $\alpha = 0$, to \emph{fairly} evaluate the two measures against each other, meaning that only dynamic node conservation is considered, or 2) $\alpha = \frac{1}{2}$, to
	give each method the \emph{best-case} advantage, since this $\alpha$ value seems to work the best for DynaWAVE \cite{vijayan2017aligning}. 
    \pgfplotsset{every tick label/.append style={font=\huge}}

\vspace{-0.3cm}

\section{Results and discussion}
	Results were gathered on a Intel i7-6700 CPU at 3.4GHz with 16GB of RAM. Execution times
	were obtained using a single core for computation. In the following tests we measure potential improvement of GoT-WAVE over DynaWAVE as follows. Let us denote by $S_G$ the (accuracy or running time) score of GoT-WAVE, and by $S_D$ the score of DynaWAVE. Also, let us denote by $G_A$ the relative gain of GoT-WAVE over DynaWAVE in terms of accuracy, and by $G_T$ the relative gain GoT-WAVE over DynaWAVE in terms of running time. Since for accuracy, a larger score is better, we define  $G_A=\frac{S_G-S_D}{min(S_G,S_D)}\times100\%$. On the other hand, since for running time, a lower score is better, we define $G_T = \frac{S_D-S_G}{min(S_G,S_D)}\times100\%$. In both cases, positive gain (i.e., a positive $G_A$ or $G_T$ value) would indicate improvement of GoT-WAVE compared to DynaWAVE, and negative gain (i.e., a negative $G_A$ or $G_T$ value) would indicate degradation of GoT-WAVE compared to DynaWAVE.  For example, in terms of accuracy, if GoT-WAVE has accuracy of 1 and DynaWAVE has accuracy of 0.7, then $G_A=\frac{1-0.7}{0.7}\times100\%=43\%$ (i.e., GoT-WAVE is superior to DynaWAVE). On the other hand, if GoT-WAVE has accuracy of  0.7 and  DynaWAVE has accuracy of 1, then $G_A=\frac{0.7-1}{0.7}\times100\%=-43\%$ (i.e., GoT-WAVE is inferior to DynaWAVE). As another example, in terms of running time, if GoT-WAVE takes 2 seconds and DynaWAVE takes 6 seconds, then $G_T=\frac{6-2}{2}\times 100\%=200\%$ (i.e., GoT-WAVE is superior to DynaWAVE). On the other hand, if  GoT-WAVE takes 6 seconds and DynaWAVE takes 2 seconds, then $G_T=\frac{2-6}{2}\times100\%=-200\%$ (i.e., GoT-WAVE is inferior to DynaWAVE).	
	
	\subsection{Evaluation using synthetic networks}\label{sec:materials}
		
	As often done~\cite{przulj2010geometric,milenkovic2010optimal,hulovatyy2015exploring}, we compare DynaWAVE and GoT-WAVE on a set of synthetic networks from different graph models. We develop temporal versions of well-known  models: \erdos~random graphs~\cite{erdos1960evolution}, \barabasi~preferential attachment~\cite{barabasi1999emergence}, \strogratz~small-world networks~\cite{watts1998collective}, geometric gene duplication model with probability cutoff~\cite{przulj2010geometric} and scale-free gene duplication~\cite{vazquez2003modeling}. A good  GPNA method should identify networks from the same model as being more topologically alike (that is, a having higher alignment quality, i.e., objective function score) than networks from different models. 
	
	\subsubsection{Synthetic networks}
	
	We generate networks with 24 snapshots each ($T = 24$). In each snapshot, new nodes arrive at the network and new edges are added to it until the desired edge density is reached. Node arrival is either linear ($N_t = \frac{N_{T}-N_1}{T-1} \cdot (t-1) + N_1$) or exponential ($N_t = N_1 \cdot e^{\frac{(t-1)}{10}}$), where $t$ is the index of the snapshot, $T$ is the total number of snapshots, $N_1$ is the number of nodes at the start and $N_t$ is the number of nodes at snapshot $t$. We set the arrival function of each model according to what was reported as the observed node arrival function for similar models~\cite{leskovec2008microscopic}. How new edges are added (i.e., which nodes they connect) is specific to each model. Edge density is set at $\approx$1\% for all models, mimicking real-world networks (such as PPIs, internet routing and email networks~\cite{melancon2006just}), and remains stable for all snapshots (e.g., this stability was observed in online social networks by \cite{hu2009evolution}). Each network starts with 100 nodes and grows to 1,000 nodes.  We generate ten networks for each of the five graph models, giving us 50 networks with 24 snapshots each, totaling to 1200 snapshots.

	\begin{table}[b]
		\caption{Results on synthetic networks when only node conservation is optimized ($\alpha=0$) or when node and edge conservation are optimized ($\alpha=\frac{1}{2}$). In parentheses, we show relative improvement (positive gain) or degradation (negative gain) in performance of GoT-WAVE compared to DynaWAVE.}
		\begin{minipage}{0.4\textwidth}
			\scriptsize
			\centering
			\label{tab:alphadynamagna}
			
			\begin{tabular}{ p{0.2cm}|l|l }
				\hline
				& \multicolumn{2}{c}{AUPR}   \\  
				$\alpha$ & \tiny \textcolor{black}{DynaWAVE} & \tiny \textcolor{black}{GoT-WAVE}  \\ \hline
				\centering 0 & 0.63 & \bf 0.79 \textcolor{black}{\tiny(+25\%)}\\ 
				$\frac{1}{2}$ & 0.59 & 0.53 \textcolor{black}{\tiny(-11\%)} \\
				
				
			\end{tabular}
			\begin{tabular}{ p{0.17cm}|l|l }
				\hline
				& \multicolumn{2}{c}{AUROC}  \\  
				$\alpha$ & \tiny \textcolor{black}{DynaWAVE} & \tiny \textcolor{black}{GoT-WAVE}  \\ \hline
				\centering 0 &  0.59 & \bf 0.78 \textcolor{black}{\tiny(+32\%)}\\ 
				\centering $\frac{1}{2}$  & 0.54 \vspace{0.03cm}&   0.70 \textcolor{black}{\tiny(+30\%)} \\
				
				\hline
				
			\end{tabular}
			
		\end{minipage}
		\begin{minipage}{0.6\textwidth}
			\scriptsize
			\centering
			\label{tab:exectimes}
			\begin{tabular}{ l|l|l  }
				\hline
				Model & \multicolumn{1}{l|}{\textcolor{black}{DGDVs}} & \multicolumn{1}{l}{\textcolor{black}{GoTs}}\\ \hline
				\textbf{Random} & 26s  & \bf 22s \textcolor{black}{\tiny(+18\%)}  \\ 
				\textbf{ScaleFree} & \bf 22s  & 25s \textcolor{black}{\tiny(-14\%)}  \\ 
				\textbf{Small-world} & 23s & \bf 4s   \textcolor{black}{\tiny(+475\%)} \\ 
				\textbf{Geo-GD} & 34s & \bf 11s   \textcolor{black}{\tiny(+210\%)} \\ 
				\textbf{ScaleFree-GD} & 16s & \bf 12s  \textcolor{black}{\tiny(+33\%)}   \\ \hline
				\multicolumn{1}{r|}{Total} & 121s &  \multicolumn{1}{c}{\bf 74s \textcolor{black}{\tiny(+64\%)}} \\
			\end{tabular}
		\end{minipage}
		
		\vspace{0.1cm}
		\begin{minipage}{0.4\textwidth}
			\centering
			(a) Accuracy.
		\end{minipage}
		\begin{minipage}{0.6\textwidth}
			\centering
			(b) Feature extraction times.
		\end{minipage}
		
	\end{table}
	
	\subsubsection{Performance on synthetic networks}\label{sec:perfsynth}
	
	With each of GoT-WAVE and DynaWAVE, we align all pairs of synthetic networks. We compute objective function scores of all alignments. We consider an objective score threshold of $k$. Then, a  pair of networks is: i) a true positive if their alignment's objective score is $k$ or higher and the networks belong to the same model, ii) a false positive if their alignment's objective score is $k$ or higher but the networks belong to different models, iii) a false negative if their alignment's objective score is lower than $k$ but the networks belong to the same model, and iv) a true negative if their alignment's objective score is lower than $k$ and the networks belong to different models. By varying $k$, we compute the area under the precision-recall (AUPR) or receiver operating characteristic (AUROC) curve. We compare GoT-WAVE and DynaWAVE with respect to these measures. Note that given the five graph models, the expected AUROC by chance is 0.2.
	
	Under the fair-case scenario, when optimizing solely node conservation ($\alpha = 0$), GoT-WAVE's AUPR and AUROC are higher by 25\%  and 32\%, respectively, than DynaWAVE's (Table~\ref{tab:alphadynamagna} (a)). 
	
	For this particular dataset, also optimizing edge conservation (i.e., $\alpha = \frac{1}{2}$) decreases performance of both methods, even though it was previously argued that  $\alpha = \frac{1}{2}$ is the best-case scenario  \cite{vijayan2017aligning,vijayan2017alignment}.  Actually, we  also verify that $\alpha = \frac{1}{2}$ is indeed the best-case scenario on our considered real networks (Section \ref{sect:real}). It is just that on our considered synthetic networks, $\alpha = 0$ happens to be both the fair and best-case scenario for both methods, and under this scenario, GoT-WAVE is superior to DynaWAVE. 
	
	On synthetic networks, we also find that extracting GoT features is overall 64\% faster than extracting DGDV features  (Table~\ref{tab:alphadynamagna} (b)). 
	%
Because both methods use their features in the same alignment strategy (WAVE), their alignment times are similar, as expected.

	
	Note that we also performed a subset of all tests for synthetic networks, and specifically those under the fair evaluation scenario ($\alpha=0$), using the other existing DGDV-based temporal GPNA method, DynaMAGNA++, and a GoT-modified version of it, which we refer to as GoT-MAGNA++. Here, we used the following values of MAGNA++'s parameters: population size of 1000 and 1000 generations. These results are qualitatively similar to those reported above: GoT-MAGNA++'s AUPR and AUROC are 16\% and 22\% higher, respectively, than DynaMAGNA++'s. However, we maybe did not give DynaMAGNA++ the best-case advantage, because this method was shown to work the best for $\alpha = \frac{1}{2}$, and under larger values of its parameters than those that we were able to consider due to MAGNA++'s high running time. So, it is possible that the performance of DynaMAGNA++ (and GoT-MAGNA++) could be improved. However, testing this would  take too much computational time, and it would do so unnecessarily, given that DynaWAVE was already shown to outperform DynaMAGNA++ in terms of both accuracy and running time on all networks but the smallest ones (with $\approx$100 nodes). So, we believe that our detailed tests against DynaWAVE are sufficient.
	
	\subsection{Evaluation using real-world networks}\label{sect:real}

	Section \ref{sec:materials} studies GPNA at the network level (whether  networks are from the same model), while here we study GPNA at the node level (whether nodes are correctly aligned). A typical process to evaluate GPNA at the node level on a real network is to insert artificial noise into the network, that is, rewire a percentage of its temporal edges (events), and align the original network to the noisy version~\cite{vijayan2017aligning}. Then, since the aligned networks have the same nodes, we can measure the percentage of all nodes that are correctly aligned.
	
	To randomize a dynamic network, we use 3 different randomization schemes. For undirected networks, we use an established randomization scheme~\cite{holme2015modern}, which we refer to as \emph{undirected randomization}. This scheme chooses two random events and swaps their time stamps with some probability. For directed networks, we use a variation of the above scheme that has an additional parameter that controls the probability of switching the edge directions of the events, which we refer to as \emph{directed randomization}. For directed networks, we use an additional randomization scheme that only swaps the edge direction of events but not their time stamps, which we refer to as \emph{pure directed randomization}. For a given scheme, we study 10 randomization (i.e., noise) levels, from 0\% to 20\% in increments of 2\%. At each noise level, we produce five random network instances and average the results over the five runs.

	\begin{table}[b]
		\scriptsize
		\centering
		\vspace{-0.2cm}
		\caption{Real temporal networks used in our experiments.}
		\label{tab:realdata}
		\begin{tabular}{ l|l|l|l|l }
			\hline
			Network  & Nodes & Events & Snaps. & Description   \\  \hline
			\textbf{zebra} \tiny \cite{rubenstein2015similar} & 27 & 500 & 57 & Zebra proximity network \\ 
			\textbf{yeast} \tiny \cite{vijayan2017alignment}  & 1,004 & 10,403 & 8 & Yeast PIN \\ 
			\textbf{aging} \tiny \cite{faisal2014dynamic}  & 6,300 & 76,666 & 38 & Human aging PIN \\ 
			\textbf{school}  \tiny\cite{gemmetto2014mitigation} & 327 & 7,388 & 5 & School proximity network \\ 
			\textbf{gallery}  \tiny\cite{isella2011s} & 420 & 22,476 & 16 & Gallery proximity network \\ 
			\textbf{arxiv}  \tiny \cite{leskovec2007graph} & 2,504 & 138,495 & 7 & Paper co-authorships \\  \hline
			
			\textbf{emails}  \tiny\cite{michalski2011matching}& 167 & 8,771 & 9 & E-mail communication\\ 
			\textbf{tennis}  \tiny\cite{aparicio2016subgraph} & 876 & 103,938 & 42 & Player dominance network \\ \hline
			
		\end{tabular}
		\vspace{-0.3cm}
	\end{table}

	First, for a given method, at each noise level, for each alignment, we compute the corresponding objective function score. Ideally, the objective score should decrease as the network is aligned to progressively noisier versions. Furthermore, since we know the perfect alignment between the original network and each of its randomized versions (as their nodes are the same), we compute the ideal objective score -- the quality of the perfect alignment, as
	measured by DynaWAVE's and GoT-WAVE's objective function. We denote the objective scores of the ideal and method-produced alignments for noise $n$ by $S_{i,n}$ and $S_{p,n}$, respectively. The expectation is that a good method's produced objective score should be similar to the method's ideal objective score, i.e., $|S_{p,n} - S_{i,n} |$ should be as close as possible to 0. Also, since we want to account for scaling (e.g., the difference of 0.1 between 0.9 and 0.8 is not the same as the difference of 0.1 between 0.3 and 0.2), we divide the difference between the produced and ideal alignment by their maximum, i.e., $max(S_{p,n}, S_{i,n})$. With these points is mind, we compute the distance $dis(S_p, S_i)$ over all considered noise levels $n$ (from 0\% to 20\%) as: $
	dis(S_p, S_i) = \sum_{n = 0\%}^{20\%} \frac{|S_{p,n} - S_{i,n} |}{max(S_{p,n}, S_{i,n})}$
	
	For each network, we compute this distance for each of GoT-WAVE and DynaWAVE. Then, we summarize gain of GoT-WAVE compared to DynaWAVE: let us denote by $S_G$ the distance score of GoT-WAVE, and by $S_D$ the score of DynaWAVE. Since a lower distance score is better, we compute the relative gain of GoT-WAVE over DynaWAVE, denoted by $G_O$, as:  $G_O=\frac{S_D-S_G}{min(S_G,S_D)}\times100\%$. Positive gains mean than GoT-WAVE is superior to DynaWAVE and negative gains mean that GoT-WAVE is inferior to DynaWAVE.
	
Second, we compare GoT-WAVE and DynaWAVE in terms of node correctness (see above). Let us denote by $S_G$ the node correctness of GoT-WAVE, and by $S_D$ the node correctness of DynaWAVE. Since higher node correctness is better, we compute the relative gain of GoT-WAVE over DynaWAVE, denoted by $G_{NC}$, as: $G_{NC}=\frac{S_G-S_D}{min(S_G,S_D)}\times100\%$. Again, positive gains mean than GoT-WAVE is superior to DynaWAVE and negative gains mean that GoT-WAVE is inferior to DynaWAVE.

	\subsubsection{Real-world temporal networks}

	We analyze eight real  networks (Table~\ref{tab:realdata}). Six of them are undirected, three of which are biological networks from the DynaWAVE study,  and three are social networks. Due to the lack of  directed biological temporal networks, we use two directed temporal networks from other fields. 

	\subsubsection{Performance on real undirected networks}

\begin{figure*}[t]
	
	\centering
	\includegraphics[width=0.6\textwidth]{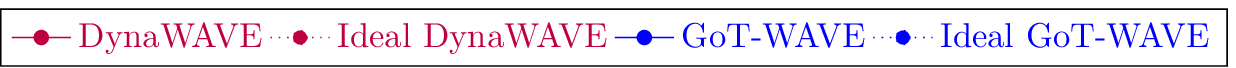}
	
	\includegraphics[width=1\textwidth]{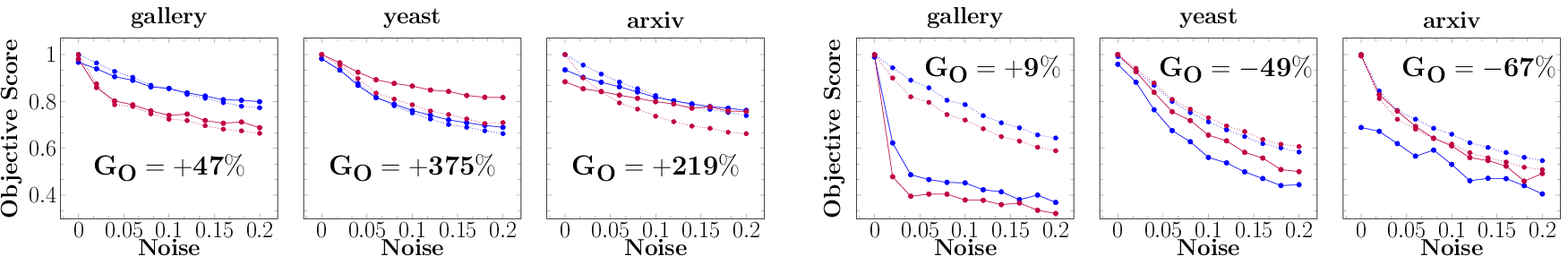}
	
	\begin{minipage}{0.49\linewidth}
		\vspace{0.1cm}
		\centering
		(a) $\alpha = 0$.
	\end{minipage}
	\begin{minipage}{0.49\linewidth}
		\vspace{0.1cm}
		\centering
		(b) $\alpha = \frac{1}{2}$.
	\end{minipage}
	
	\caption{Comparison between GoT-WAVE and DynaWAVE on undirected networks in terms of how well their alignments' objective scores match the objective scores of ideal alignments, when (a) only node conservation is optimized ($\alpha=0$) and (b) both node and edge conservation are optimized ($\alpha=\frac{1}{2}$). Recall that $G_O$ is the relative gain of GoT-WAVE over DynaWAVE (positive: GoT-WAVE is superior; negative: DynaWAVE is superior). 
	}
	\label{fig:undirscore}
	
\end{figure*}

	In terms of the objective score (Figure~\ref{fig:undirscore} both DynaWAVE and GoT-WAVE show adequate behavior, i.e., the objective score decreases as we add more noise. When optimizing solely node conservation ($\alpha = 0$), we observe that: (i)  for \textbf{gallery} and \textbf{zebra} networks, both methods closely match their ideal alignments over all noise levels; (ii) for \textbf{yeast} and \textbf{aging} networks, both methods closely match their ideal alignments for low noise levels, but for high noises levels, DynaWAVE drifts away from its ideal alignments while GoT-WAVE still closely matches its ideal alignments; and (iii) for \textbf{arxiv} and \textbf{school} networks, both methods are far from their ideal alignments for low noise levels, but for high noise levels, GoT-WAVE closely matches its ideal alignments while DynaWAVE is still far from its ideal alignments. In other words, in terms of the total gain $G_O$, GoT-WAVE improves upon DynaWAVE, more closely matches its ideal alignments than DynaWAVE, for all six networks. When optimizing both node and edge conservation ($\alpha = \frac{1}{2}$),  GoT-WAVE more closely matches its ideal alignments for two out of the six networks (\textbf{gallery} and \textbf{aging}). So, the two methods can be seen as complementary.
	
\begin{table}[b]
	\scriptsize
	\centering
	\caption{Node correctness when aligning an undirected real network to itself (noise = 0). In parentheses, we show relative improvement (positive gain) or degradation
		(negative gain) in performance of GoT-WAVE compared to DynaWAVE. In bold, we show the best result for each network.}
	\label{tab:dynawavereal_correctness}
	\begin{tabular}{ l|l|l||l|l  }
		\multicolumn{1}{c}{} & \multicolumn{2}{c}{(a) $\alpha = 0$} & \multicolumn{2}{c}{(b) $\alpha = \frac{1}{2}$} 
		\vspace{0.1cm}\\
		\hline
		Network & \textcolor{black}{DynaWAVE} & \textcolor{black}{GoT-WAVE} & \textcolor{black}{DynaWAVE} & \textcolor{black}{GoT-WAVE}\\ \hline
		\textbf{zebra} & \bf 0.926 $\pm$ 0.05  &  0.578 $\pm$ 0.09 \textcolor{black}{\tiny(-60\%)} & 0.911 $\pm$ 0.04  & 0.615 $\pm$ 0.14  \textcolor{black}{\tiny(-48\%)} \\ 
		\textbf{yeast} & \bf0.966 $\pm$ 0.01 & 0.924 $\pm$ 0.01 \textcolor{black}{\tiny(-5\%)} & \bf 0.966 $\pm$ 0.01 &  0.919 $\pm$ 0.01 \textcolor{black}{\tiny(-5\%)}\\ 
		\textbf{aging} & 0.912 $\pm$ 0.01 & 0.942 $\pm$ 0.01  \textcolor{black}{\tiny(+3\%)} & \bf 0.959 $\pm$ 0.01 &  0.955 $\pm$ 0.01 \textcolor{black}{\tiny(-0.4\%)}\\ 
		\textbf{arxiv} & 0.340 $\pm$ 0.02 & 0.446 $\pm$ 0.02  \textcolor{black}{\tiny(+31\%)} & \bf0.658 $\pm$ 0.01 &  0.602 $\pm$ 0.04 \textcolor{black}{\tiny(-9\%)}\\ 
		\textbf{gallery} & 0.507 $\pm$ 0.03 & 0.485 $\pm$ 0.03  \textcolor{black}{\tiny(-5\%)} & \bf 0.557 $\pm$ 0.01 &  0.531 $\pm$ 0.01 \textcolor{black}{\tiny(-5\%)}\\ 
		\textbf{school} & 0.735 $\pm$ 0.03  &  0.861 $\pm$ 0.03 \textcolor{black}{\tiny(+17\%)} & \bf 0.973 $\pm$ 0.01 &  0.971 +- 0.01 \textcolor{black}{\tiny(-0.2\%)}\\  \hline
		
	\end{tabular}
\end{table}
	
\begin{figure*}[t]
	\centering
	\includegraphics[width=0.28\textwidth]{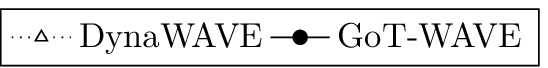}
	
	\includegraphics[width=1\textwidth]{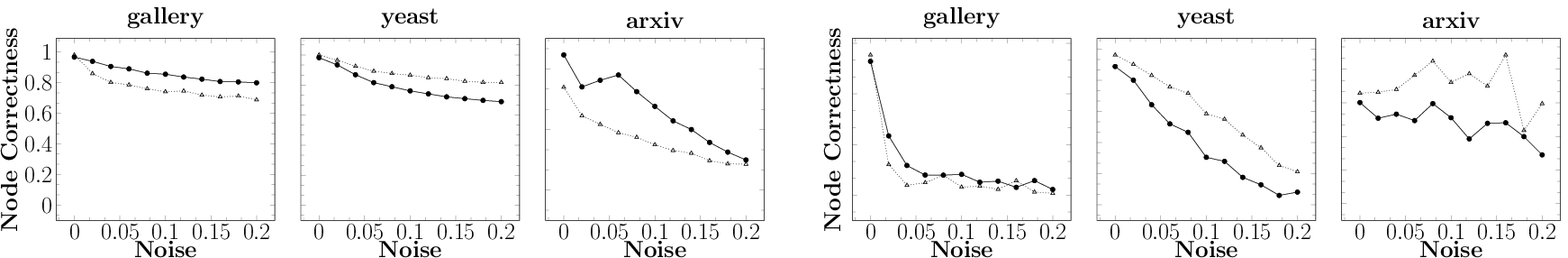}
	
	\begin{minipage}{0.49\linewidth}
		\vspace{0.1cm}
		\centering
		(a) $\alpha = 0$.
	\end{minipage}
	\begin{minipage}{0.49\linewidth}
		\vspace{0.1cm}
		\centering
		(b) $\alpha = \frac{1}{2}$.
	\end{minipage}
	
	\caption{Comparison between GoT-WAVE and DynaWAVE on undirected networks in terms of node correctness, when (a) only node conservation is optimized ($\alpha=0$) and (b) both node and edge conservation are optimized ($\alpha=\frac{1}{2}$). The higher the node correctness, the better the method.
	}
	\label{fig:undirnc}
	
\end{figure*}

	In terms of node correctness (Table~\ref{tab:dynawavereal_correctness}, Figure \ref{fig:undirnc}), for $\alpha = 0$, the two methods are again complementary - each is the best for three of the six networks. For $\alpha = \frac{1}{2}$, DynaWAVE's node correctness improves more substantially than GoT-WAVE's, which is why now DynaWAVE is superior for most (though not all) of the networks. In short, GoT-WAVE already captures some of the information that DWEC captures and thus does not benefit much from using it, while DynaWAVE captures different information from DWEC and thus benefits more from  using it. 
	Note that the superiority of one method over the other one is typically consistent over all noise levels, for both $\alpha = 0$ and $\alpha = \frac{1}{2}$.

	In terms of running time, extracting GoT features is faster than extracting DGDV features for the sparser networks (\textbf{zebra}, \textbf{aging} and \textbf{school}) and slower for the denser networks (\textbf{aging}, \textbf{arxiv} and \textbf{gallery}). Denser networks induce more GoTs than dynamic graphlets, and thus, GoTs are computationally heavier. Just as  for synthetic networks, because both methods use the same alignment strategy (WAVE), their alignment times are similar.

	\subsubsection{Performance on real directed networks}

	\begin{table}[b]
		
		\caption{Node correctness when aligning a directed network to itself (noise = 0), for $\alpha = 0$.  In parentheses, we show relative improvement (positive gain) or degradation
			(negative gain) in performance of GoT-WAVE compared to DynaWAVE. Node correctness results over all noise levels, using the best GoT-WAVE version (3-node directed GoTs), are shown in Figure~\ref{fig:dirncscore} (a) and (b) for $\alpha = 0$ and $\alpha = \frac{1}{2}$, respectively.
		}
		\label{tab:dynawavereal_correctness_dir}
		\small
		\begin{tabular}{ l|l|l|l|l  }
			\hline
			\multirow{2}{*}{Network} & \multicolumn{1}{c|}{\textcolor{black}{DynaWAVE}} & \multicolumn{3}{c}{\textcolor{black}{GoT-WAVE}}\\
			&  \textcolor{black}{Undirected-4} & \textcolor{black}{Undirected-4} & \textcolor{black}{Directed-3} & \textcolor{black}{Directed-4}   \\  \hline
			\textbf{emails} & \bf 0.85 $\pm$ 0.02 & 0.81 $\pm$ 0.03  & 0.83 $\pm$ 0.01 \textcolor{black}{(-2\%)} &  0.81 $\pm$ 0.02\\ 
			\textbf{tennis} &  0.74 $\pm$ 0.01  & 0.84 $\pm$ 0.03 & \bf 0.85 $\pm$  0.02 \textcolor{black}{(+15\%)} &  0.81 $\pm$ 0.02  \\  \hline
			
		\end{tabular}
	\end{table}
	
	As expected, since this scheme only rewires edge direction, the original network and the noisy networks have identical topology when ignoring edge directions. Because of this, and because DGDVs are undirected, DynaWAVE can not differentiate between the networks, while GoT-WAVE can, since GoTs accounts for edge directions. The rest of this section focuses on the other, directed randomization scheme, where not only edge directions but also time stamps are rewired.
	
	Unlike previous sections, we first address node correctness and only then objective score. We choose this organization because, on directed networks, we do experiments with different sets of GoTs (i.e., 4-node undirected GoTs, 3-node directed GoTs, and 4-node directed GoTs) in an effort to find the best set. For simplicity, we choose the best GoTs as those with the highest node correctness when aligning the original network to a noiseless version for $\alpha = 0$. We find that 3-node directed GoTs are the best for both of the directed networks (Table~\ref{tab:dynawavereal_correctness_dir}). Thus, henceforth, we use 3-node directed GoTs (for DGDVs, we still use four nodes and six events, as recommended by the DGDV authors).
	
			\begin{figure*}[t]
		\begin{minipage}{0.49\textwidth}
			\centering
			\includegraphics[width=0.28\textwidth]{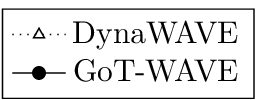}
		\end{minipage}	
		\begin{minipage}{0.49\textwidth}
			\centering
			\includegraphics[width=0.6\textwidth]{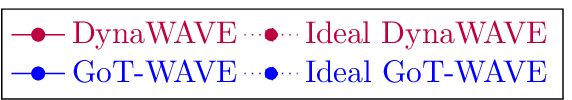}
		\end{minipage}	
		\vspace{0.1cm}
		
		\begin{minipage}{0.35\textwidth}
			\centering
			
			\includegraphics[width=1\textwidth]{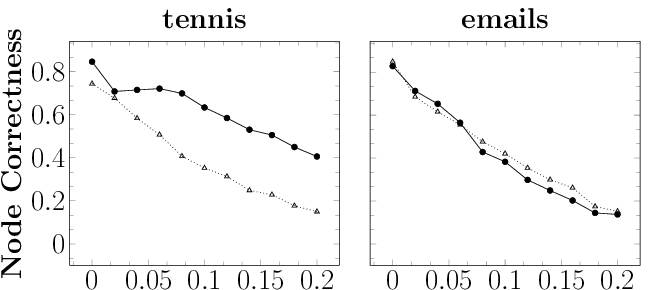}

			\includegraphics[width=1\textwidth]{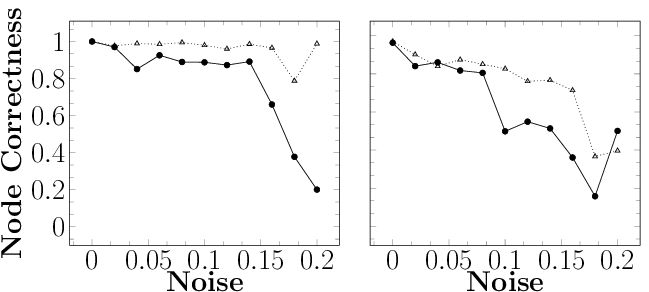}
			
		\end{minipage}
		\begin{minipage}{0.07\textwidth}
			\centering
			
			(a) \\ $\alpha = 0$
			
			\vspace{0.6cm}
			
			(b) \\ $\alpha = \frac{1}{2}$
			
		\end{minipage}
		\hspace{0.8cm}
		\begin{minipage}{0.35\textwidth}
			\centering
			
			\includegraphics[width=1\textwidth]{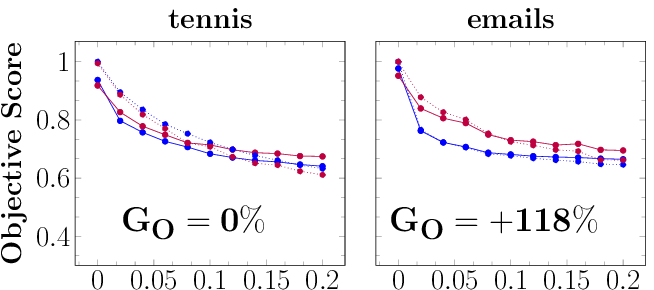}

			\includegraphics[width=1\textwidth]{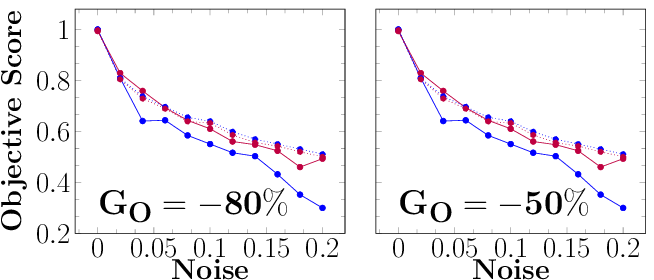}
			
		\end{minipage}
		\begin{minipage}{0.07\textwidth}
			\centering
			
			(c) \\ $\alpha = 0$
			
			\vspace{0.6cm}
			
			(d) \\ $\alpha = \frac{1}{2}$
			
		\end{minipage}

		\caption{Comparison between GoT-WAVE and DynaWAVE on directed networks in terms of (a,b) node correctness and (c,d) how well their alignments' objective scores match the objective scores of ideal alignments, when (a,c) only node conservation is optimized ($\alpha=0$) and (b,d) both node and edge conservation are optimized ($\alpha=\frac{1}{2}$). For panels (a,b), the higher the node correctness value, the better the method. For panels (c,d), recall that $G_O$ is the relative gain of GoT-WAVE over DynaWAVE (positive: GoT-WAVE is superior; negative: DynaWAVE is superior). } 
		\label{fig:dirncscore}
		
		\vspace{-0.2cm}
	\end{figure*}
	
	In terms of node correctness, for $\alpha = 0$, we observe that GoT-WAVE has higher correctness than DynaWAVE for \textbf{tennis} over noise levels, and overall comparable node correctness for \textbf{emails}, depending on the noise level (Figure~\ref{fig:dirncscore} (a)). We hypothesize that GoT-WAVE's performance depends on subgraph overlap between (consecutive) snapshots of the input  network. Subgraph overlap is expected to be higher in the \textbf{tennis} network than in the \textbf{emails} network, because tennis players tend to have the same opponents every year, while one might not necessarily email the same people in different time periods. Indeed, these are exactly the trends that our two networks show. The same trend was already observed for an alternative email network \cite{crawford2018cluenet}.  Networks with low subgraph overlap such as our \textbf{emails} network have fewer transitions (i.e., lower GoTs frequencies), and thus provide less information to GoT-WAVE. 
	For $\alpha = \frac{1}{2}$, DynaWAVE's node correctness is higher for both networks  over most noise levels (Figure~\ref{fig:dirncscore}~(b)).
	
	In terms of the objective score, for $\alpha = 0$, GoT-WAVE more closely matches its ideal alignments than DynaWAVE does for \textbf{emails}, and the two are comparable for \textbf{tennis} (Figure~\ref{fig:dirncscore} (c)). For \textbf{tennis}, GoT-WAVE mismatches the ideal alignments at lower noise levels but matches them at higher noise levels, while  DynaWAVE mismatches the ideal alignments at both lower and higher noise level. For $\alpha = \frac{1}{2}$, DynaWAVE's performance is again better for both networks (Figure~\ref{fig:dirncscore} (d)). 
	
In terms of running time, results are qualitatively similar to those for undirected networks.

    \section{Conclusion}
	
    We propose GoT-WAVE as a new algorithm for temporal GPNA.
	Our results suggest that GoTs
	are an efficient measure of dynamic node conservation. While DynaWAVE benefits
	more from also optimizing dynamic edge conservation, only GoT-WAVE can support directed edges. Future work on
	better incorporating dynamic edge conservation into GoT-WAVE may yield further improvements. Also,   GoTs could be used under newer alignment strategies instead of WAVE. Further, on real networks, each of GoTs and DGDVs is superior half the time and the two dynamic node conservation measures are thus complementary. So, a deep understanding of each measure's (dis)advantages  could perhaps guide development of a new, improved measure. As more temporal real data continue to become available, which is  inevitable, dynamic network analyses, including temporal GPNA, will continue to gain importance.

	\section*{Funding}
	
	This work is partially funded by FCT (Portuguese Foundation
	for Science and Technology) within project UID/EEA/50014/2013 and
	by the United States AFOSR (Air Force Office of Scientific Research)
	[YIP FA9550-16-1-0147]. David Apar\'icio is supported by a FCT/MAP-i PhD research grant
	[PD/BD/105801/2014].

	\bibliographystyle{acm}
	\bibliography{refs}
	
\end{document}


\title{Supplementary material for \\ GoT-WAVE: Temporal network alignment \\ using graphlet-orbit transitions}
\author{D. Apar\'icio$^1$, P. Ribeiro$^1$, T. Milenkovi\'c$^2$ and F. Silva$^1$ \vspace{0.1cm} \\ 
$^1$ CRACS \& INESC-TEC, and the Department of Computer Science, \\Faculty of Sciences of the University of Porto, \\ Porto, 4169-007,
Portugal\vspace{0.1cm}\\
$^2$ Department of Computer Science and Engineering, ECK Institute  for Global Health,  \\ and Interdisciplinary Center for Network Science
and Applications (iCeNSA), \\ University of Notre Dame, Notre Dame, IN 46556, USA. \vspace{0.1cm}\\
E-mail: daparicio@dcc.fc.up.pt}

\date{}
\maketitle

\section{Extracting Graphlet-orbit Transitions (GoTs)}\label{sup_sec:extract_features}

Supplementary Algorithm~S\ref{sup_alg:network_sim} describes how we obtain $k$-node GoTs of a given network. First, our method generates all $k$-node graphlets and their respective orbits and populates a g-trie with them (lines 2 and 3). A g-trie is a datastructure that efficiently stores input subgraphs and greatly speeds up their enumeration in a network~\cite{ribeiro2014}. Note that our method generates the g-trie (for a given set of orbits) only once, stores it in a file and reuses it when necessary. The enumeration process is repeated for all $T$ snapshots, storing the enumerated occurrences in a vector of the form of $\{Nodes_{\{1..k\}}, t, Orbits_{\{1..k\}}\}$. We use this representation so that, after we sort the vector (line 10), the occurrences of the same subgraph $Nodes_{\{1..k\}}$ are in subsequent positions of vector $Occs$ and sorted by $t$. We compare each consecutive pair of occurrences $(Occ_1, Occ_2)$ and, if they belong to the same subgraph (i.e., the nodes are the same), we update the $GoTs$ of each node according to its orbit transition from $t$ to $t+k$ (lines 11-14).

\begin{supalgorithm}
	\caption{Obtain all $k$-node GoTs of network $N$ with $T$ snapshots}
	\label{sup_alg:network_sim}
	\begin{algorithmic}[1]
		\Procedure{enumerateGoT($N$, $T$, $k$)}{}
		\State $\mathcal{O}_k :$ generate all $k$-node orbits.
		\State $GT_k = \Call{createGTrie}{\mathcal{O}_k}$ 
		\State $GoTs: $ $Nodes(N) \times |\mathcal{O}_k| \times |\mathcal{O}_k|$ tensor.
		\State
		\ForAll {$t \in \{1:T\}$}
		\State $\mathcal{S}_{N,t} :$ snapshot $t$ of $N$
		\State $Occs\{t\} = \Call{enumerateOrbitOccurrences}{\mathcal{S}_{N,t}, GT_k}$ 
		\EndFor
		\State
		\State \Call{sort}{$Occs$}
		\ForAll {$(Occ_1, Occ_2) \in Occs$}
		\If{$Nodes(Occ_1) == Nodes(Occ_2)$}
		\ForAll {$i \in \{1:k\}$}
		\State $GoTs[Node_i(Occ)][Orbit_i(Occ_1), Orbit_i(Occ_2)]$ ++
		\EndFor
		\EndIf
		\EndFor
		\State \Return $GoTs$
		\EndProcedure
		
	\end{algorithmic}
\end{supalgorithm}

\section{Temporal random graph models}\label{sup_sec:random_models}

Following the discussion from the main paper, we give specific details on the models. Remember that: at each snapshot $t$, new nodes arrive in the network and new edges are added until the desired edge density is reached (1\%). Which edges are connected depends on the specifics of the model. Supplementary Table~S\ref{sup_tab:models} summarizes the models.

\begin{itemize}
	
	\setlength{\itemsep}{3pt}
	\item \textbf{\erdos~(Random):} Two nodes are chosen at random and connected. Past edges are kept.
	\item \textbf{\barabasi~(ScaleFree):} Two nodes $u$ and $v$ are chosen at random but they become connected only if $\frac{max(deg(u), deg(v))}{|E|} > r$, where $r$ is a randomly generated value $\in [0, 1]$. Therefore, nodes with higher degrees have a bigger chance of gaining new connections ("the rich get richer"). Past edges are kept.
	\item \textbf{\strogratz~(Small-world):} At $t=0$ an initial ring network is created, where each node is connected to $\approx k$ neighbors. Since $N_0 = 100$ and the edge density is 1\%, $k=1$ for $t=0$. Edges are then randomly rewired with probability $\beta = 0.2$. For $t>0$, new nodes arrive and a ring is formed again (with a possibly different $k$) while keeping the rewired connections previously added. Rewiring is again performed, both on the new ring and on the old randomized edges. Therefore, past edges might disappear.
	\item \textbf{Geometric gene duplication~(Geo-GD):} Initially $k$ seed-nodes are placed close-by in a two-dimensional space: $d(u,v)^2 < \epsilon$. As suggested by \cite{przulj2010geometric}, $\epsilon = 5\cdot10^{-2}$ and $k=5$ are used. Nodes are incrementally added to the network one at a time, and each new node $u$ is placed at a distance $d(u,f)^2 \leq \epsilon$ from a random \textit{father-node} $f$ already in the network. The model includes parameter $p$ which controls how likely node $u$ is to \textit{cut-off} from $f$; for our purposes $p=0.2$ since it gives origin to realistic PPI networks~\cite{przulj2010geometric}. When a node cuts-off from its father, it is placed at a distance of $10\epsilon$ at most. Node additions stop when the desired number of nodes is achieved and the closest 1$\%$ edges are added to snapshot $t$, while the other possible edges remain unactivated. It is possible that past edges disappear since close edges in snapshot $t$ are not guaranteed to remain in the 1\% closest edges of snapshot $t+1$.
	\item \textbf{Scale-free gene duplication~(ScaleFree-GD):} After a few seed edges are added to the network, each new node is (i) connected to a random father-node and it (ii) copies the father's connections. The model has two parameters: $p$ controls the likelihood of child- and father-nodes being connected by an edge, and $q$ sets the probability of the child-node keeping his father's connections to other nodes. For our purposes, $p = 0.3$ and $q = 0.7$ since these values generated realistic PPI networks~\cite{przulj2010geometric}. The model also sets a 50\% chance that when an edge is not successfully replicated from father to child either (a) the father keeps the edge but the child does not copy it or (b) the child steals the connection from the father; therefore, past edges can be lost.
\end{itemize}

\pagebreak

\section{Precision-recall curves and ROC curves}\label{sup_sec:pr_curves}

With each of GoT-WAVE and DynaWAVE, we align all pairs of synthetic
networks. We compute objective function scores of all alignments. Objective scores vary from 0 (i.e., the networks are completerly different) to 1 (i.e., the networks are exactly alike). We
consider an objective score threshold of $k$, also varying from 0 to 1. Then, a pair of networks is: i)
a true positive if their alignment's objective score is $k$ or higher and the
networks belong to the same model, ii) a false positive if their alignment’s
objective score is $k$ or higher but the networks belong to different models,
iii) a false negative if their alignment's objective score is lower than $k$ but
the networks belong to the same model, and iv) a true negative if their
alignment's objective score is lower than $k$ and the networks belong to
different models. 

More formally, if $G$ are $H$ are two networks being aligned, $c(X)$ is the class (i.e., model) of the network and $Score(G, H)$ is their objective score, then, for a given threshold $k$:

\[
Label(G, H, k) = 
\begin{cases} 
\textnormal{True Positive} & Score(G, H) \geq k, c(G) = c(H) \\
\textnormal{False Positive} & Score(G, H) \geq k, c(G) \not= c(H) \\
\textnormal{True Negative} & Score(G, H) < k, c(G) \not= c(H) \\
\textnormal{False Negative} & Score(G, H) < k, c(G) = c(H) 
\end{cases}
\]

Then, for each level $k$, we compute precision, recall (also called true positive rate (TPR)) and false negative rate (FPR).

\begin{equation}
Pr(k) = \frac{|\textnormal{True Positives}|}{|\textnormal{True Positives} \cup \textnormal{False Positives}|} \tag{Precision}
\end{equation}

\begin{equation}
Rec(k) / TPR(k) = \frac{|\textnormal{True Positives}|}{|\textnormal{True Positives} \cup \textnormal{False Negatives}|} \tag{Recall/TPR}
\end{equation}

\begin{equation}
FPR(k) = \frac{|\textnormal{False Positives}|}{|\textnormal{False Positives} \cup {\textnormal{True Negatives}|} \tag{FPR}}
\end{equation}

We vary $k$ in increments of $0.01$, resulting in 1000 pairs of Precision versus Recall and, similarly, in 1000 pairs of TPR versus FPR. We then compute the area under the precision-recall (AUPR) or receiver operating characteristic (AUROC) curve as follows:

\begin{equation}\label{eq:prrecall}
AUPR = \sum_{k=0.01}^{1} Pr(k) \Delta Rec(k)
\end{equation}

\begin{equation}\label{eq:auroc}
AUROC = \sum_{k=0.01}^{1} TPR(k) \Delta FPR(k)
\end{equation}

\pagebreak

\section{Randomization schemes}\label{sup_section:schemes}

\subsection{Undirected randomization}

This randomization scheme was proposed in~\cite{holme2015modern} and used in~\cite{vijayan2017alignment,vijayan2017aligning}. Given the original undirected dynamic network $\mathcal{S}_N$ a randomization percentage $p$, one randomly picks edge $e_1$ to be rewired. We then pick another random edge $e_2$ and, with probability $p$, we rewire the two events. That is, given $e_1 = (u, v, S_i, S_f)$ and $e_2 = (u', v', S_i', S_f')$ (where $S_i$ and $S_i'$ are the starting snapshots, and $S_f$ and $S_f'$ are the ending snapshots) we do one of the following transformations with 50\% probability:

\begin{itemize}
	\item $e_1 = (u, v, S_i, S_f)$ $\rightarrow$ $e_1 = (u, v', S_i, S_f)$ and \\
	$e_2 = (u', v', S_i', S_f')$ $\rightarrow$ $e_2 = (u', v, S_i, S_f)$, or
	\item $e_1 = (u, v, S_i, S_f)$ $\rightarrow$ $e_1 = (u, u', S_i, S_f)$ and \\
	$e_2 = (u', v', S_i', S_f')$ $\rightarrow$ $e_2 = (v, v', S_i, S_f)$, or
\end{itemize}

If the transformation was performed, $e_1$ and $e_2$ are both taken out of the list of edges to be rewired. Otherwise, only $e_1$ is taken out. The process is followed until no edges are to be rewired.

\subsection{Directed randomization}

This randomization scheme is adapted from~\cite{holme2015modern} to directed networks. Given the original directed dynamic network $\mathcal{S}_N$ a randomization percentage $p$, one randomly picks edge $e_1$ to be rewired. We then pick another random edge $e_2$ and, with probability $p$, we rewire the two events. That is, given $e_1 = (u, v, S_i, S_f)$ and $e_2 = (u', v', S_i', S_f')$ (where $S_i$ and $S_i'$ are the starting snapshots, and $S_f$ and $S_f'$ are the ending snapshots) we do one of the following transformations with 50\% probability:

\begin{itemize}
	\item $e_1 = (u, v, S_i, S_f)$ $\rightarrow$ $e_1 = (u, v', S_i, S_f)$ and \\
	$e_2 = (u', v', S_i', S_f')$ $\rightarrow$ $e_2 = (u', v, S_i, S_f)$, or
	\item $e_1 = (u, v, S_i, S_f)$ $\rightarrow$ $e_1 = (u, u', S_i, S_f)$ and \\
	$e_2 = (u', v', S_i', S_f')$ $\rightarrow$ $e_2 = (v, v', S_i, S_f)$, or
\end{itemize}

If the transformation was performed, an additional parameter $\gamma$ controls edge reversal. So, with probability $\gamma$, one performs the transformation for each edge:

\begin{itemize}
	\item $e_k = (x, y, S_n, S_m)$ $\rightarrow$ $e_k = (y, x, S_n, S_m)$
\end{itemize}

Then, $e_1$ and $e_2$ are both taken out of the list of edges to be rewired. Otherwise, only $e_1$ is taken out. The process is followed until no edges are to be rewired.

\subsection{Pure directed randomization}

Given the original dynamic directed network $\mathcal{S}_N$, one randomly picks edge $e_1$ to be rewired. That is, given $e_1 = (u, v, S_i, S_f)$ (where $S_i$ is the starting snapshot, and $S_f$ is the ending snapshot), and a randomization percentage $p$, we the following transformations with $p\%$ probability:

\begin{itemize}
	\item $e_1 = (u, v, S_i, S_f)$ $\rightarrow$ $e_1 = (v, u, S_i, S_f)$
\end{itemize}

Otherwise, $e_1$ is kept as it was. $e_1$ is taken out of the list of edges to be rewired and the process is followed until no edges are to be rewired.

\pgfplotstableread{../plots/dirstrict_dg_wave_emails_noise_score.dat}{\dirstrictdgnoiseemailsscore}
\pgfplotstableread{../plots/dirstrict_idealdg_wave_emails_noise_score.dat}{\dirstrictidealdgnoiseemailsscore}
\pgfplotstableread{../plots/dirstrict_got_wave_emails_noise_score.dat}{\dirstrictgotnoiseemailsscore}
\pgfplotstableread{../plots/dirstrict_idealgot_wave_emails_noise_score.dat}{\dirstrictidealgotnoiseemailsscore}

\pgfplotstableread{../plots/dirstrict_dg_wave_tennis_noise_score.dat}{\dirstrictdgnoisetennisscore}
\pgfplotstableread{../plots/dirstrict_idealdg_wave_tennis_noise_score.dat}{\dirstrictidealdgnoisetennisscore}
\pgfplotstableread{../plots/dirstrict_got_wave_tennis_noise_score.dat}{\dirstrictgotnoisetennisscore}
\pgfplotstableread{../plots/dirstrict_idealgot_wave_tennis_noise_score.dat}{\dirstrictidealgotnoisetennisscore}

\pgfplotstableread{../plots/dirstrict_dg_wave_california_noise_score.dat}{\dirstrictdgnoisecaliforniascore}
\pgfplotstableread{../plots/dirstrict_idealdg_wave_california_noise_score.dat}{\dirstrictidealdgnoisecaliforniascore}
\pgfplotstableread{../plots/dirstrict_got_wave_california_noise_score.dat}{\dirstrictgotnoisecaliforniascore}
\pgfplotstableread{../plots/dirstrict_idealgot_wave_california_noise_score.dat}{\dirstrictidealgotnoisecaliforniascore}

\pagebreak

\section*{Supplementary Tables}

\begin{suptable}[h!]
	\centering
	
	\caption{Set of graph models used in our experiments. All networks (regardless of the model) have 24 snapshots, start with 100 nodes, grow until they reach 1000 nodes, and have edge density of $= 1\%$ in all snapshots. Node arrival (linear or exponential) is set to what was reported in~\cite{leskovec2008microscopic} for similar models. How nodes are connected (i.e., how new edges are added) depends on the model (see Supplementary Section~S\ref{sup_sec:random_models}).}
	\vspace{0.3cm}
	\begin{tabular}{l| l |l} 
		\hline
		Model &  Node arrival &  New edges \\ \hline
		
		\textbf{Random} & linear & random  \\
		
		\textbf{ScaleFree} & exponential & preferential Attachment \\
		
		\textbf{Small-world} & linear & ring + rewire \scriptsize ($\beta$ = $0.2$)\\
		
		\textbf{Geo-GD} & linear & duplication w/ cut-off \scriptsize ($p$ = $0.2$)\\
		
		\textbf{ScaleFree-GD} & exponential & duplication w/ edge loss \scriptsize ($p$ = $0.3$, $q$ = $0.7$)\\ \hline
		
	\end{tabular}
	
	\label{sup_tab:models}
	
\end{suptable}

\begin{suptable}[!h]
	\centering
	\caption{Average time to align two networks when using DGDVs or GoTs. We compute the time of aligning each model (e.g., the time to align each of the \textbf{Random} networks with any other networks). Since each of the 5 models $m$ has 10 instances $n$ (i.e., networks), we consider $\frac{(n-1)\times n}{2} = 45$ alignments of a given network to networks of its own model (e.g., align two \textbf{Random} networks) and $n \times (m-1)\times n = 400$ alignments to networks of different models (e.g., align a \textbf{Random} network with a \textbf{ScaleFree} network). For each model, we average the time over all 445 considered alignments.}
	
	\vspace{0.3cm}
	\label{sub_tab:aligntime}
	\begin{tabular}{ l|l|l  }
		\hline
		Model & \multicolumn{1}{c|}{DGDVs} & \multicolumn{1}{c}{GoTs}\\ \hline
		\bf	Random &  6.219s &  \textbf{6.078s \footnotesize (+2\%)}\\ 
		\bf	ScaleFree & 6.196s & \textbf{6.056s \footnotesize (+2\%)} \\ 
		\bf	Small-world &  6.181s & \textbf{6.049s \footnotesize (+2\%)} \\ 
		\bf	Geo-GD & 6.027s & \textbf{5.925s \footnotesize (+2\%)}\\ 
		\bf	ScaleFree-GD &  6.009s  & \textbf{5.868s \footnotesize (+2\%)}\\ \hline
		\multicolumn{1}{r|}{Total} &  30.632s  & \textbf{29.976s \footnotesize (+2\%)} 
		
	\end{tabular}
\end{suptable}

\begin{suptable}[!h]
	
	\caption{Results on undirected real networks in terms of (a) feature extraction times and (b) number occurrences (i.e., number of dynamic graphlets or GoTs found on the network). GoTs induce many more occurrences than DGDVs, and this is especially true for denser networks, such as \textbf{aging}, \textbf{arxiv} and \textbf{school}. Due to our fast enumeration algorithm based on g-tries~\cite{ribeiro2014}, GoTs' extraction is still faster for the sparser networks, namely \textbf{zebra}, \textbf{yeast} and \textbf{school}. In parentheses, we show relative improvement (positive gain) or degradation
		(negative gain) in performance of GoT-WAVE compared to DynaWAVE. We assume that more occurrences means degradation. Thus, GoT-WAVE shows a much higher degradation in terms of number of occurrences than in execution time (-1,886\% versus -142\%), showcasing our enumeration algorithm's efficiency.
	}
	\label{sup_table:real_times}
	\vspace{0.3cm}
	\centering
	\begin{tabular}{l|l|l|l|l}
		\multicolumn{1}{c}{}& \multicolumn{2}{c}{(a)} & \multicolumn{2}{c}{(b)} \vspace{0.1cm} \\ 
		\hline
		\multirow{2}{*}{Network} & \multicolumn{2}{c|}{Execution time} & \multicolumn{2}{c}{\#Occurrences}\\ 
		 & DGDVs & GoTs & DGDVs & GoTs\\ \hline
		\bf zebra & 0.06s & \textbf{0.02s \footnotesize (+200\%)} & 
		$9.676 \times 10^3$ & $\mathbf{7.792 \times 10^3}$ \textbf{\footnotesize (+24\%)} \\
		\bf yeast  & 86s & \textbf{65s \footnotesize (+32\%)} & $\mathbf{7.160\times10^6}$ & $5.815\times10^7$ \footnotesize (-712\%)\\
		\bf aging  & \textbf{202s} & 696s \footnotesize (-245\%) & $\mathbf{1.532\times10^7}$ & $5.510\times10^8$ \footnotesize (-3,496\%)\\
		\bf school  & 4s & \textbf{2s \footnotesize (+100\%)} & $\mathbf{3.765\times10^5}$ & $2.352\times10^6$ \footnotesize (-525\%) \\
		\bf arxiv  & \textbf{586s} & 1,360s \footnotesize (-132\%) & $\mathbf{6.178\times10^7}$ & $1.067\times10^9$ \footnotesize (-1,627\%)  \\
		\bf gallery &  \textbf{6s} & 14s \footnotesize (-133\%) & $\mathbf{5.864\times10^5}$ & $1.432\times10^7$ \footnotesize (-2,341\%) \\ \hline
		\multicolumn{1}{r|}{Total} & \bf 884s & 2,137s \footnotesize (-142\%) & $\mathbf{8.523\times10^7}$ &  $1.693\times10^9$ \footnotesize (-1,886\%)
	\end{tabular}
\end{suptable}

\begin{suptable}[!h]
	\centering
	\caption{Average time to align two networks when using DGDVs or GoTs. We compute the time of aligning a network with each of its randomized versions. Since each network randomization has ten noise levels, each comprised of five networks, we consider a total of 50 alignments per network, and average out the times.}
	
	\vspace{0.3cm}
	\label{sup_table:real_align}
	\begin{tabular}{ l|l|l  }
		\hline
		Network & \multicolumn{1}{c|}{DGDVs} & \multicolumn{1}{c}{GoTs}\\ \hline
		\bf	zebra & \bf 4.50s  & 5.36s \footnotesize (-19\%)\\ 
		\bf	yeast & \bf 55.32s & 82.46s \footnotesize (-49\%) \\ 
		\bf	aging & 2,264.92s &  \bf 1,209.58s \footnotesize (+87\%) \\ 
		\bf	arxiv & 416.40s & \bf 249.98s \footnotesize (+67\%)\\ 
		\bf	gallery & 20.36s & \bf19.70s \footnotesize (+3\%)\\ 
		\bf	school & 20.46s & \bf14.72s \footnotesize (+39\%)\\ \hline
		\multicolumn{1}{r|}{Total} &  2,781.95s  & \bf 1,586.79\footnotesize (75\%) 
		
	\end{tabular}
\end{suptable}

\begin{suptable}[!h]
	\caption{Results on directed real networks in terms of feature extraction times when using DGDVs or GoTs. For DGDVs, we extract dynamic graphlets with up to four nodes and up to six events, as suggested in~\cite{hulovatyy2015exploring}. For GoTs, we extract undirected GoTs with up to four nodes, directed GoTs with up to three nodes, and directed GoTs with up to four nodes. In parentheses, we show relative improvement (positive gain)
		or degradation (negative gain) in performance of GoT-WAVE compared to
		DynaWAVE. In bold, we show the best result for each network.}
	\label{sup_table:dir_times}
	\vspace{0.3cm}
	\centering
	\begin{tabular}{l|l|l|l|l}
		\hline
		\multirow{2}{*}{Network}& \multicolumn{1}{c|}{DGDVs} & \multicolumn{3}{c}{GoTs} \\ 
		& Undirected-4 & Undirected-4 & Directed-3 & Directed-4 \\ \hline
		\bf tennis & 29.09s & 113.07s \footnotesize (-289\%) & \bf 2.90s \footnotesize (+903\%) & 128.43s \footnotesize (-341\%)\\
		\bf emails  & 5.19s & 5.65s \footnotesize (-9\%) & \bf 0.21s \footnotesize (+2,371\%) & 7.84s \footnotesize (-51\%)\\ \hline
		\multicolumn{1}{r|}{Total} & 34.28s & 118.72s \footnotesize (-246\%) & \bf 3.11s \footnotesize (+1,002\%)& 136.27s \footnotesize (-297\%)
	\end{tabular}
\end{suptable}

\clearpage

\section*{Supplementary Figures}
\pgfplotsset{every tick label/.append style={font=\huge}}

\begin{figure*}[!h]
	\centering
	\begin{tikzpicture}[scale=0.8]
	\begin{axis}[%
	legend columns=-1, 
	hide axis,
	xmin=1,
	xmax=1,
	ymin=0,
	ymax=0.4,
	]
	\addlegendimage{purple,mark=*}
	\addlegendentry{\textcolor{purple}{DynaWAVE}};
	\addlegendimage{purple,mark=*, dotted}
	\addlegendentry{\textcolor{purple}{Ideal DynaWAVE}};
	\addlegendimage{blue,mark=*}
	\addlegendentry{\textcolor{blue}{GoT-WAVE}};
	\addlegendimage{blue,mark=*, dotted}
	\addlegendentry{\textcolor{blue}{Ideal GoT-WAVE}};
	\end{axis}
	\end{tikzpicture}
	\vspace{-3.9cm}
	
	\begin{tikzpicture}[scale=0.33]
	\begin{axis}[minor tick num=2,
	xlabel=\textbf{Noise},
	ymin=0.3,
	ylabel near ticks,
	xmin=-0.02,
	xmax=0.22,
	title=\textbf{\huge gallery},
	xticklabel style={
		/pgf/number format/fixed,
		/pgf/number format/precision=5
	},
	scaled x ticks=false,
	title style={font=\large},
	label style={font=\huge},
	ylabel=\textbf{Objective Score}],
	\addplot [mark=*, blue] table [x={noise}, y={score}] {../new_plots/score_alpha0_gallery_got_.dat};
	\addplot [mark=*,dotted, blue] table [x={noise}, y={score}] {../new_plots/score_alpha0_gallery_idealgot_.dat};
	\addplot [mark=*, purple] table [x={noise}, y={score}] {../new_plots/score_alpha0_gallery_dg_.dat};
	\addplot [mark=*,dotted, purple] table [x={noise}, y={score}] {../new_plots/score_alpha0_gallery_idealdg_.dat};
	
	\node[color=black] 
	at (axis cs:0.1,0.52) {\Huge  $\mathbf{G_O = +47\%}$};
	\end{axis}
	\end{tikzpicture}
	\begin{tikzpicture}[scale=0.33]
	\begin{axis}[minor tick num=2,
	xlabel=\textbf{Noise},
	ymin=0.3,
	ylabel near ticks,
	xmin=-0.02,
	xmax=0.22,
	title=\textbf{\huge yeast},
	xticklabel style={
		/pgf/number format/fixed,
		/pgf/number format/precision=5
	},
	yticklabels={,,},
	scaled x ticks=false,
	title style={font=\large},
	label style={font=\huge},
	ylabel=\textbf{}],
	\addplot [mark=*, blue] table [x={noise}, y={score}] {../new_plots/score_alpha0_yeast_got_.dat};
	\addplot [mark=*,dotted, blue] table [x={noise}, y={score}] {../new_plots/score_alpha0_yeast_idealgot_.dat};
	\addplot [mark=*, purple] table [x={noise}, y={score}] {../new_plots/score_alpha0_yeast_dg_.dat};
	\addplot [mark=*,dotted, purple] table [x={noise}, y={score}] {../new_plots/score_alpha0_yeast_idealdg_.dat};

	\node[color=black] 
	at (axis cs:0.1,0.52) {\Huge  $\mathbf{G_O = +375\%}$};
	\end{axis}
	\end{tikzpicture}
	\begin{tikzpicture}[scale=0.33]
	\begin{axis}[minor tick num=2,
	xlabel=\textbf{Noise},
	ymin=0.3,
	ylabel near ticks,
	xmin=-0.02,
	xmax=0.22,
	title=\textbf{\huge arxiv},
	xticklabel style={
		/pgf/number format/fixed,
		/pgf/number format/precision=5
	},
	yticklabels={,,},
	scaled x ticks=false,
	title style={font=\large},
	label style={font=\huge},
	ylabel=\textbf{}],
	\addplot [mark=*, blue] table [x={noise}, y={score}] {../new_plots/score_alpha0_arxiv_got_.dat};
	\addplot [mark=*,dotted, blue] table [x={noise}, y={score}] {../new_plots/score_alpha0_arxiv_idealgot_.dat};
	\addplot [mark=*, purple] table [x={noise}, y={score}] {../new_plots/score_alpha0_arxiv_dg_.dat};
	\addplot [mark=*,dotted, purple] table [x={noise}, y={score}] {../new_plots/score_alpha0_arxiv_idealdg_.dat};
	
	\node[color=black] 
	at (axis cs:0.1,0.52) {\Huge  $\mathbf{G_O = +219\%}$};
	\end{axis}
	\end{tikzpicture}
	\begin{minipage}{0.02\textwidth}
		\textcolor{white}{.}
	\end{minipage}
	\begin{tikzpicture}[scale=0.33]
	\begin{axis}[minor tick num=2,
	xlabel=\textbf{Noise},
	ymin=0.3,
	ylabel near ticks,
	xmin=-0.02,
	xmax=0.22,
	title=\textbf{\huge gallery},
	xticklabel style={
		/pgf/number format/fixed,
		/pgf/number format/precision=5
	},
	yticklabels={,,},
	scaled x ticks=false,
	title style={font=\large},
	label style={font=\huge},
	ylabel=\textbf{Objective Score}],
	\addplot [mark=*, blue] table [x={noise}, y={score}] {../new_plots/score_alpha0.5_gallery_got_.dat};
	\addplot [mark=*,dotted, blue] table [x={noise}, y={score}] {../new_plots/score_alpha0.5_gallery_idealgot_.dat};
	\addplot [mark=*, purple] table [x={noise}, y={score}] {../new_plots/score_alpha0.5_gallery_dg_.dat};
	\addplot [mark=*,dotted, purple] table [x={noise}, y={score}] {../new_plots/score_alpha0.5_gallery_idealdg_.dat};

	\node[color=black] 
	at (axis cs:0.13,0.94) {\Huge  $\mathbf{G_O = +9\%}$};
	\end{axis}
	\end{tikzpicture}
	\begin{tikzpicture}[scale=0.33]
	\begin{axis}[minor tick num=2,
	xlabel=\textbf{Noise},
	ymin=0.3,
	ylabel near ticks,
	xmin=-0.02,
	xmax=0.22,
	title=\textbf{\huge yeast},
	xticklabel style={
		/pgf/number format/fixed,
		/pgf/number format/precision=5
	},
	yticklabels={,,},
	scaled x ticks=false,
	title style={font=\large},
	label style={font=\huge},
	ylabel=\textbf{}],
	\addplot [mark=*, blue] table [x={noise}, y={score}] {../new_plots/score_alpha0.5_yeast_got_.dat};
	\addplot [mark=*,dotted, blue] table [x={noise}, y={score}] {../new_plots/score_alpha0.5_yeast_idealgot_.dat};
	\addplot [mark=*, purple] table [x={noise}, y={score}] {../new_plots/score_alpha0.5_yeast_dg_.dat};
	\addplot [mark=*,dotted, purple] table [x={noise}, y={score}] {../new_plots/score_alpha0.5_yeast_idealdg_.dat};
	
	\node[color=black] 
	at (axis cs:0.12,0.94) {\Huge  $\mathbf{G_O = -49\%}$};
	\end{axis}
	\end{tikzpicture}
	\begin{tikzpicture}[scale=0.33]
	\begin{axis}[minor tick num=2,
	xlabel=\textbf{Noise},
	ymin=0.3,
	ylabel near ticks,
	xmin=-0.02,
	xmax=0.22,
	title=\textbf{\huge arxiv},
	xticklabel style={
		/pgf/number format/fixed,
		/pgf/number format/precision=5
	},
	yticklabels={,,},
	scaled x ticks=false,
	title style={font=\large},
	label style={font=\huge},
	ylabel=\textbf{}],
	\addplot [mark=*, blue] table [x={noise}, y={score}] {../new_plots/score_alpha0.5_arxiv_got_.dat};
	\addplot [mark=*,dotted, blue] table [x={noise}, y={score}] {../new_plots/score_alpha0.5_arxiv_idealgot_.dat};
	\addplot [mark=*, purple] table [x={noise}, y={score}] {../new_plots/score_alpha0.5_arxiv_dg_.dat};
	\addplot [mark=*,dotted, purple] table [x={noise}, y={score}] {../new_plots/score_alpha0.5_arxiv_idealdg_.dat};
	
	\node[color=black] 
	at (axis cs:0.12,0.94) {\Huge  $\mathbf{G_O = -67\%}$};
	\end{axis}
	\end{tikzpicture}
	
	\begin{tikzpicture}[scale=0.33]
	\begin{axis}[minor tick num=2,
	xlabel=\textbf{Noise},
	ymin=0.3,
	ylabel near ticks,
	xmin=-0.02,
	xmax=0.22,
	title=\textbf{\huge zebra},
	xticklabel style={
		/pgf/number format/fixed,
		/pgf/number format/precision=5
	},
	scaled x ticks=false,
	title style={font=\large},
	label style={font=\huge},
	ylabel=\textbf{Objective Score}],
	\addplot [mark=*, blue] table [x={noise}, y={score}] {../new_plots/score_alpha0_zebra_got_.dat};
	\addplot [mark=*,dotted, blue] table [x={noise}, y={score}] {../new_plots/score_alpha0_zebra_idealgot_.dat};
	\addplot [mark=*, purple] table [x={noise}, y={score}] {../new_plots/score_alpha0_zebra_dg_.dat};
	\addplot [mark=*,dotted, purple] table [x={noise}, y={score}] {../new_plots/score_alpha0_zebra_idealdg_.dat};
	
	\node[color=black] 
	at (axis cs:0.1,0.52) {\Huge  $\mathbf{G_O = +3\%}$};
	\end{axis}
	\end{tikzpicture}
	\begin{tikzpicture}[scale=0.33]
	\begin{axis}[minor tick num=2,
	xlabel=\textbf{Noise},
	ymin=0.3,
	ylabel near ticks,
	xmin=-0.02,
	xmax=0.22,
	title=\textbf{\huge aging},
	xticklabel style={
		/pgf/number format/fixed,
		/pgf/number format/precision=5
	},
	yticklabels={,,},
	scaled x ticks=false,
	title style={font=\large},
	label style={font=\huge},
	ylabel=\textbf{}],
	\addplot [mark=*, blue] table [x={noise}, y={score}] {../new_plots/score_alpha0_aging_got_.dat};
	\addplot [mark=*,dotted, blue] table [x={noise}, y={score}] {../new_plots/score_alpha0_aging_idealgot_.dat};
	\addplot [mark=*, purple] table [x={noise}, y={score}] {../new_plots/score_alpha0_aging_dg_.dat};
	\addplot [mark=*,dotted, purple] table [x={noise}, y={score}] {../new_plots/score_alpha0_aging_idealdg_.dat};

	\node[color=black] 
	at (axis cs:0.1,0.52) {\Huge  $\mathbf{G_O = +73\%}$};
	\end{axis}
	\end{tikzpicture}
	\begin{tikzpicture}[scale=0.33]
	\begin{axis}[minor tick num=2,
	xlabel=\textbf{Noise},
	ymin=0.3,
	ylabel near ticks,
	xmin=-0.02,
	xmax=0.22,
	title=\textbf{\huge school},
	xticklabel style={
		/pgf/number format/fixed,
		/pgf/number format/precision=5
	},
	yticklabels={,,},
	scaled x ticks=false,
	title style={font=\large},
	label style={font=\huge},
	ylabel=\textbf{}],
	\addplot [mark=*, blue] table [x={noise}, y={score}] {../new_plots/score_alpha0_school_got_.dat};
	\addplot [mark=*,dotted, blue] table [x={noise}, y={score}] {../new_plots/score_alpha0_school_idealgot_.dat};
	\addplot [mark=*, purple] table [x={noise}, y={score}] {../new_plots/score_alpha0_school_dg_.dat};
	\addplot [mark=*,dotted, purple] table [x={noise}, y={score}] {../new_plots/score_alpha0_school_idealdg_.dat};
	
	\node[color=black] 
	at (axis cs:0.1,0.52) {\Huge  $\mathbf{G_O = +11\%}$};
	\end{axis}
	\end{tikzpicture}
	\begin{minipage}{0.02\textwidth}
		\textcolor{white}{.}
	\end{minipage}
	\begin{tikzpicture}[scale=0.33]
	\begin{axis}[minor tick num=2,
	xlabel=\textbf{Noise},
	ymin=0.3,
	ylabel near ticks,
	xmin=-0.02,
	xmax=0.22,
	title=\textbf{\huge zebra},
	xticklabel style={
		/pgf/number format/fixed,
		/pgf/number format/precision=5
	},
	yticklabels={,,},
	scaled x ticks=false,
	title style={font=\large},
	label style={font=\huge},
	ylabel=\textbf{Objective Score}],
	\addplot [mark=*, blue] table [x={noise}, y={score}] {../new_plots/score_alpha0.5_zebra_got_.dat};
	\addplot [mark=*,dotted, blue] table [x={noise}, y={score}] {../new_plots/score_alpha0.5_zebra_idealgot_.dat};
	\addplot [mark=*, purple] table [x={noise}, y={score}] {../new_plots/score_alpha0.5_zebra_dg_.dat};
	\addplot [mark=*,dotted, purple] table [x={noise}, y={score}] {../new_plots/score_alpha0.5_zebra_idealdg_.dat};

	\node[color=black] 
	at (axis cs:0.12,0.94) {\Huge  $\mathbf{G_O = -39\%}$};
	\end{axis}
	\end{tikzpicture}
	\begin{tikzpicture}[scale=0.33]
	\begin{axis}[minor tick num=2,
	xlabel=\textbf{Noise},
	ymin=0.3,
	ylabel near ticks,
	xmin=-0.02,
	xmax=0.22,
	title=\textbf{\huge aging},
	xticklabel style={
		/pgf/number format/fixed,
		/pgf/number format/precision=5
	},
	yticklabels={,,},
	scaled x ticks=false,
	title style={font=\large},
	label style={font=\huge},
	ylabel=\textbf{}],
	\addplot [mark=*, blue] table [x={noise}, y={score}] {../new_plots/score_alpha0.5_aging_got_.dat};
	\addplot [mark=*,dotted, blue] table [x={noise}, y={score}] {../new_plots/score_alpha0.5_aging_idealgot_.dat};
	\addplot [mark=*, purple] table [x={noise}, y={score}] {../new_plots/score_alpha0.5_aging_dg_.dat};
	\addplot [mark=*,dotted, purple] table [x={noise}, y={score}] {../new_plots/score_alpha0.5_aging_idealdg_.dat};
	
	\node[color=black] 
	at (axis cs:0.12,0.94) {\Huge  $\mathbf{G_O = +14\%}$};
	\end{axis}
	\end{tikzpicture}
	\begin{tikzpicture}[scale=0.33]
	\begin{axis}[minor tick num=2,
	xlabel=\textbf{Noise},
	ymin=0.3,
	ylabel near ticks,
	xmin=-0.02,
	xmax=0.22,
	title=\textbf{\huge school},
	xticklabel style={
		/pgf/number format/fixed,
		/pgf/number format/precision=5
	},
	yticklabels={,,},
	scaled x ticks=false,
	title style={font=\large},
	label style={font=\huge},
	ylabel=\textbf{}],
	\addplot [mark=*, blue] table [x={noise}, y={score}] {../new_plots/score_alpha0.5_school_got_.dat};
	\addplot [mark=*,dotted, blue] table [x={noise}, y={score}] {../new_plots/score_alpha0.5_school_idealgot_.dat};
	\addplot [mark=*, purple] table [x={noise}, y={score}] {../new_plots/score_alpha0.5_school_dg_.dat};
	\addplot [mark=*,dotted, purple] table [x={noise}, y={score}] {../new_plots/score_alpha0.5_school_idealdg_.dat};
	
	\node[color=black] 
	at (axis cs:0.065,0.41) {\Huge  $\mathbf{G_O = -36\%}$};
	\end{axis}
	\end{tikzpicture}

	\begin{minipage}{0.49\linewidth}
		\vspace{0.1cm}
		\centering
		(a) $\alpha = 0$.
	\end{minipage}
	\begin{minipage}{0.49\linewidth}
		\vspace{0.1cm}
		\centering
		(a) $\alpha = \frac{1}{2}$.
	\end{minipage}
	
	\caption{Comparison between GoT-WAVE and DynaWAVE on undirected networks in terms of how well their alignments' objective scores match the objective scores of ideal alignments, when (a) only node conservation is optimized ($\alpha=0$) and (b) both node and edge conservation are optimized ($\alpha=\frac{1}{2}$). Recall that $G_O$ is the relative gain of GoT-WAVE over DynaWAVE (positive: GoT-WAVE is superior; negative: DynaWAVE is superior).
	}\label{sub_fig:undirscore}

\end{figure*}

\begin{figure*}[!h]
	\centering
	\begin{tikzpicture} 
	\begin{axis}[%
	legend columns=-1, 
	hide axis,
	xmin=1,
	xmax=1,
	ymin=0,
	ymax=0.4,
	]
	\addlegendimage{black,mark=triangle, dotted, mark options={solid}}
	\addlegendentry{\textcolor{black}{DynaWAVE}};
	\addlegendimage{black,mark=*}
	\addlegendentry{\textcolor{black}{GoT-WAVE}};
	\end{axis}
	\end{tikzpicture}
	\vspace{-4.9cm}
	
	\begin{tikzpicture}[scale=0.33]
	\begin{axis}[minor tick num=2,
	xlabel=\textbf{Noise},
	ymin=-0.1,
	ylabel near ticks,
	xmin=-0.02,
	xmax=0.22,
	title=\textbf{\huge gallery},
	xticklabel style={
		/pgf/number format/fixed,
		/pgf/number format/precision=5
	},
	scaled x ticks=false,
	title style={font=\large},
	label style={font=\huge},
	ylabel=\textbf{Node Correctness}],
	\addplot [mark=*, black,] table [x={noise}, y={score}] {../new_plots/nodecorr_alpha0.0_gallery_got_.dat};
	\addplot [mark=*, black, dotted] table [x={noise}, y={score}] {../new_plots/nodecorr_alpha0.0_gallery_dg_.dat};
	
	\end{axis}
	\end{tikzpicture}
	\begin{tikzpicture}[scale=0.33]
	\begin{axis}[minor tick num=2,
	xlabel=\textbf{Noise},
	ymin=-0.1,
	ylabel near ticks,
	xmin=-0.02,
	xmax=0.22,
	title=\textbf{\huge yeast},
	xticklabel style={
		/pgf/number format/fixed,
		/pgf/number format/precision=5
	},
	yticklabels={,,},
	scaled x ticks=false,
	title style={font=\large},
	label style={font=\huge},
	ylabel=\textbf{}],
	\addplot [mark=*, black] table [x={noise}, y={score}] {../new_plots/nodecorr_alpha0.0_yeast_got_.dat};
	\addplot [mark=*, black, dotted] table [x={noise}, y={score}] {../new_plots/nodecorr_alpha0.0_yeast_dg_.dat};

	\end{axis}
	\end{tikzpicture}
	\begin{tikzpicture}[scale=0.33]
	\begin{axis}[minor tick num=2,
	xlabel=\textbf{Noise},
	ymin=-0.1,
	ylabel near ticks,
	xmin=-0.02,
	xmax=0.22,
	title=\textbf{\huge arxiv},
	xticklabel style={
		/pgf/number format/fixed,
		/pgf/number format/precision=5
	},
	yticklabels={,,},
	scaled x ticks=false,
	title style={font=\large},
	label style={font=\huge},
	ylabel=\textbf{}],
	\addplot [mark=*, black] table [x={noise}, y={score}] {../new_plots/nodecorr_alpha0.0_arxiv_got_.dat};
	\addplot [mark=*, black, dotted] table [x={noise}, y={score}] {../new_plots/nodecorr_alpha0.0_arxiv_dg_.dat};
	
	\end{axis}
	\end{tikzpicture}
	\begin{minipage}{0.02\textwidth}
		\textcolor{white}{.}
	\end{minipage}
	\begin{tikzpicture}[scale=0.33]
	\begin{axis}[minor tick num=2,
	xlabel=\textbf{Noise},
	ymin=-0.1,
	ylabel near ticks,
	xmin=-0.02,
	xmax=0.22,
	title=\textbf{\huge gallery},
	xticklabel style={
		/pgf/number format/fixed,
		/pgf/number format/precision=5
	},
	yticklabels={,,},
	scaled x ticks=false,
	title style={font=\large},
	label style={font=\huge},
	ylabel=\textbf{Node Correctness}],
	\addplot [mark=*, black] table [x={noise}, y={score}] {../new_plots/nodecorr_alpha0.5_gallery_got_.dat};
	\addplot [mark=*, black, dotted] table [x={noise}, y={score}] {../new_plots/nodecorr_alpha0.5_gallery_dg_.dat};

	\end{axis}
	\end{tikzpicture}
	\begin{tikzpicture}[scale=0.33]
	\begin{axis}[minor tick num=2,
	xlabel=\textbf{Noise},
	ymin=-0.1,
	ylabel near ticks,
	xmin=-0.02,
	xmax=0.22,
	title=\textbf{\huge yeast},
	xticklabel style={
		/pgf/number format/fixed,
		/pgf/number format/precision=5
	},
	yticklabels={,,},
	scaled x ticks=false,
	title style={font=\large},
	label style={font=\huge},
	ylabel=\textbf{}],
	\addplot [mark=*, black] table [x={noise}, y={score}] {../new_plots/nodecorr_alpha0.5_yeast_got_.dat};
	\addplot [mark=*, black, dotted] table [x={noise}, y={score}] {../new_plots/nodecorr_alpha0.5_yeast_dg_.dat};
	
	\end{axis}
	\end{tikzpicture}
	\begin{tikzpicture}[scale=0.33]
	\begin{axis}[minor tick num=2,
	xlabel=\textbf{Noise},
	ymin=-0.1,
	ylabel near ticks,
	xmin=-0.02,
	xmax=0.22,
	title=\textbf{\huge arxiv},
	xticklabel style={
		/pgf/number format/fixed,
		/pgf/number format/precision=5
	},
	yticklabels={,,},
	scaled x ticks=false,
	title style={font=\large},
	label style={font=\huge},
	ylabel=\textbf{}],
	\addplot [mark=*, black] table [x={noise}, y={score}] {../new_plots/nodecorr_alpha0.5_arxiv_got_.dat};
	\addplot [mark=*, black, dotted] table [x={noise}, y={score}] {../new_plots/nodecorr_alpha0.5_arxiv_dg_.dat};
	
	\end{axis}
	\end{tikzpicture}
	\begin{tikzpicture}[scale=0.33]
	\begin{axis}[minor tick num=2,
	xlabel=\textbf{Noise},
	ymin=-0.1,
	ylabel near ticks,
	xmin=-0.02,
	xmax=0.22,
	title=\textbf{\huge zebra},
	xticklabel style={
		/pgf/number format/fixed,
		/pgf/number format/precision=5
	},
	scaled x ticks=false,
	title style={font=\large},
	label style={font=\huge},
	ylabel=\textbf{Node Correctness}],
	\addplot [mark=*, black] table [x={noise}, y={score}] {../new_plots/nodecorr_alpha0_zebra_got_.dat};
	\addplot [mark=*, black, dotted] table [x={noise}, y={score}] {../new_plots/nodecorr_alpha0_zebra_dg_.dat};
	
	\end{axis}
	\end{tikzpicture}
	\begin{tikzpicture}[scale=0.33]
	\begin{axis}[minor tick num=2,
	xlabel=\textbf{Noise},
	ymin=-0.1,
	ylabel near ticks,
	xmin=-0.02,
	xmax=0.22,
	title=\textbf{\huge aging},
	xticklabel style={
		/pgf/number format/fixed,
		/pgf/number format/precision=5
	},
	yticklabels={,,},
	scaled x ticks=false,
	title style={font=\large},
	label style={font=\huge},
	ylabel=\textbf{}],
	\addplot [mark=*, black] table [x={noise}, y={score}] {../new_plots/nodecorr_alpha0_aging_got_.dat};
	\addplot [mark=*, black, dotted] table [x={noise}, y={score}] {../new_plots/nodecorr_alpha0_aging_dg_.dat};

	\end{axis}
	\end{tikzpicture}
	\begin{tikzpicture}[scale=0.33]
	\begin{axis}[minor tick num=2,
	xlabel=\textbf{Noise},
	ymin=-0.1,
	ylabel near ticks,
	xmin=-0.02,
	xmax=0.22,
	title=\textbf{\huge school},
	xticklabel style={
		/pgf/number format/fixed,
		/pgf/number format/precision=5
	},
	yticklabels={,,},
	scaled x ticks=false,
	title style={font=\large},
	label style={font=\huge},
	ylabel=\textbf{}],
	\addplot [mark=*, black] table [x={noise}, y={score}] {../new_plots/nodecorr_alpha0_school_got_.dat};
	\addplot [mark=*, black, dotted] table [x={noise}, y={score}] {../new_plots/nodecorr_alpha0_school_dg_.dat};
	
	\end{axis}
	\end{tikzpicture}
	\begin{minipage}{0.02\textwidth}
		\textcolor{white}{.}
	\end{minipage}
	\begin{tikzpicture}[scale=0.33]
	\begin{axis}[minor tick num=2,
	xlabel=\textbf{Noise},
	ymin=-0.1,
	ylabel near ticks,
	xmin=-0.02,
	xmax=0.22,
	title=\textbf{\huge zebra},
	xticklabel style={
		/pgf/number format/fixed,
		/pgf/number format/precision=5
	},
	yticklabels={,,},
	scaled x ticks=false,
	title style={font=\large},
	label style={font=\huge},
	ylabel=\textbf{Node Correctness}],
	\addplot [mark=*, black] table [x={noise}, y={score}] {../new_plots/nodecorr_alpha0.5_zebra_got_.dat};
	\addplot [mark=*, black, dotted] table [x={noise}, y={score}] {../new_plots/nodecorr_alpha0.5_zebra_dg_.dat};

	\end{axis}
	\end{tikzpicture}
	\begin{tikzpicture}[scale=0.33]
	\begin{axis}[minor tick num=2,
	xlabel=\textbf{Noise},
	ymin=-0.1,
	ylabel near ticks,
	xmin=-0.02,
	xmax=0.22,
	title=\textbf{\huge aging},
	xticklabel style={
		/pgf/number format/fixed,
		/pgf/number format/precision=5
	},
	yticklabels={,,},
	scaled x ticks=false,
	title style={font=\large},
	label style={font=\huge},
	ylabel=\textbf{}],
	\addplot [mark=*, black] table [x={noise}, y={score}] {../new_plots/nodecorr_alpha0.5_aging_got_.dat};
	\addplot [mark=*, black, dotted] table [x={noise}, y={score}] {../new_plots/nodecorr_alpha0.5_aging_dg_.dat};
	
	\end{axis}
	\end{tikzpicture}
	\begin{tikzpicture}[scale=0.33]
	\begin{axis}[minor tick num=2,
	xlabel=\textbf{Noise},
	ymin=-0.1,
	ylabel near ticks,
	xmin=-0.02,
	xmax=0.22,
	title=\textbf{\huge school},
	xticklabel style={
		/pgf/number format/fixed,
		/pgf/number format/precision=5
	},
	yticklabels={,,},
	scaled x ticks=false,
	title style={font=\large},
	label style={font=\huge},
	ylabel=\textbf{}],
	\addplot [mark=*, black] table [x={noise}, y={score}] {../new_plots/nodecorr_alpha0.5_school_got_.dat};
	\addplot [mark=*, black, dotted] table [x={noise}, y={score}] {../new_plots/nodecorr_alpha0.5_school_dg_.dat};
	
	\end{axis}
	\end{tikzpicture}
	
	\caption{Comparison between GoT-WAVE and DynaWAVE on undirected networks in terms of node correctness, when (a) only node conservation is optimized ($\alpha=0$) and (b) both node and edge conservation are optimized ($\alpha=\frac{1}{2}$). The higher the node correctness, the better the method. 
	}
	\label{sup_fig:undirnodecorr}
	
\end{figure*}

\begin{supfigure}[!h]
	\includegraphics[width=1\linewidth]{example.eps}
	\caption{In the main paper we observe that, when both node and edge conservation are considered ($\alpha = \frac{1}{2}$), in contrast to when only node conservation is considered ($\alpha = 0$), DynaWAVE has  better results than GoT-WAVE consistently. Here we hypothesize why that might the case with an example. When we use GoTs, node $a$ from network $G$ and node $a'$ from network $H$ are correctly identified as different (i.e., node $a$ and node $a'$ have different GoTs). When we use DGDVs, node $a$ and node $a'$ are incorrectly identified as similar/equal (i.e., node $a$ and node $a'$ have the same DGDVs). For details on DGDVs we refer the reader to~\cite{hulovatyy2015exploring}. DGDVs can not distinguish between nodes $a$ and $a'$ because a dynamic graphlet only allows for one event per time-step. For instance, the last graphlet in the DGDVs is not allowed (and not calculated). Thus, for this case, DGDVs can only distinguish nodes $a$ and $a'$ when edge conservation is also considered. Cases such as these show why DGDVs have bigger benefits in using DWEC than GoTs.
	}
	\label{sup_fig:example}
\end{supfigure}

\begin{figure*}[!h]
	\centering
	\begin{tikzpicture}[scale=0.8]
	\begin{axis}[%
	legend columns=-1, 
	hide axis,
	xmin=1,
	xmax=1,
	ymin=0,
	ymax=0.4,
	]
	\addlegendimage{purple,mark=*}
	\addlegendentry{\textcolor{purple}{DynaWAVE}};
	\addlegendimage{purple,mark=*, dotted}
	\addlegendentry{\textcolor{purple}{Ideal DynaWAVE}};
	\addlegendimage{blue,mark=*}
	\addlegendentry{\textcolor{blue}{GoT-WAVE}};
	\addlegendimage{blue,mark=*, dotted}
	\addlegendentry{\textcolor{blue}{Ideal GoT-WAVE}};
	\end{axis}
	\end{tikzpicture}
	\vspace{-3.9cm}

	\begin{minipage}{0.32\textwidth}
		\begin{tikzpicture}[scale=0.5]
		\begin{axis}[minor tick num=2,
		xlabel=\textbf{Noise},
		ymin=0.3,
		ylabel near ticks,
		ymax=1.05,
		xmin=-0.02,
		xmax=0.22,
		title=\textbf{\huge emails},
		xticklabel style={
			/pgf/number format/fixed,
			/pgf/number format/precision=5
		},
		scaled x ticks=false,
		title style={font=\Huge},
		label style={font=\Large},
		ylabel=\textbf{Objective Score}],
		\addplot [mark=*, purple] table [x={noise}, y={score}] {\dirstrictdgnoiseemailsscore};
		\addplot [mark=*, blue] table [x={noise}, y={score}] {\dirstrictgotnoiseemailsscore};
		\addplot [mark=triangle, purple,dotted, mark options={solid}] table [x={noise}, y={score}] {\dirstrictidealdgnoiseemailsscore};
		\addplot [mark=triangle, blue,dotted, mark options={solid}] table [x={noise}, y={score}] {\dirstrictidealgotnoiseemailsscore};

		\end{axis}
		\end{tikzpicture}
	\end{minipage}
	\begin{minipage}{0.32\textwidth}
		\begin{tikzpicture}[scale=0.5]
		\begin{axis}[minor tick num=2,
		xlabel=\textbf{Noise},
		ymin=0.3,
		ylabel near ticks,
		ymax=1.05,
		xmin=-0.02,
		xmax=0.22,
		title=\textbf{\huge tennis},
		xticklabel style={
			/pgf/number format/fixed,
			/pgf/number format/precision=5
		},
		scaled x ticks=false,
		title style={font=\Huge},
		label style={font=\Large},
		ylabel=\textbf{Objective Score}],
		\addplot [mark=*, purple] table [x={noise}, y={score}] {\dirstrictdgnoisetennisscore};
		\addplot [mark=*, blue] table [x={noise}, y={score}] {\dirstrictgotnoisetennisscore};
		\addplot [mark=triangle, purple,dotted, mark options={solid}] table [x={noise}, y={score}] {\dirstrictidealdgnoisetennisscore};
		\addplot [mark=triangle, blue,dotted, mark options={solid}] table [x={noise}, y={score}] {\dirstrictidealgotnoisetennisscore};

		\end{axis}
		\end{tikzpicture}
	\end{minipage}

	\caption{Comparison between GoT-WAVE and DynaWAVE on directed networks in terms of how well their alignments' objective scores match the objective scores of ideal alignments, when only node conservation is optimized ($\alpha=0$). The noisy versions are generated using the pure directed randomization scheme (i.e., time stamps of two events are never swapped, only the edge direction is swapped). We observe, on one hand, as expected, that DynaWAVE does not distinguish between the original and its randomized versions because DGDVs do not take edge direction into account. On the other hand, GoT-WAVE clearly distinguishes between the original and its randomized versions because GoTs take edge direction into account.
	}
	\label{sup_figure:dirstuff}
\end{figure*}

\clearpage

\section*{Funding}

This material is based upon work partially funded by FCT (Portuguese Foundation
for Science and Technology) within project UID/EEA/50014/2013 and
by the United States AFOSR (Air Force Office of Scientific Research)
[YIP FA9550-16-1-0147]. David Apar\'icio is supported by a FCT/MAP-i PhD research grant
[PD/BD/105801/2014]. 

\bibliographystyle{acm}
\bibliography{refs}   